\documentclass[11pt,a4paper]{article}
\usepackage{times,latexsym}
\usepackage{url}
\usepackage{adjustbox}
\usepackage[T1]{fontenc}
\usepackage{subcaption}
\usepackage{algorithm}
\usepackage{algorithmicx}
\usepackage{xcolor}
\usepackage{graphicx}
\usepackage{amsmath}
\usepackage{dsfont}
\usepackage{pgfplots}

\usepackage[acceptedWithA]{tacl2018v2}

\usepackage{xspace,mfirstuc,tabulary}

\newcommand{\com}[1]{}

\newif\iftaclinstructions
\taclinstructionsfalse 
\iftaclinstructions
\renewcommand{\confidential}{}
\renewcommand{\anonsubtext}{(No author info supplied here, for consistency with
TACL-submission anonymization requirements)}
\newcommand{\instr}
\fi

\iftaclpubformat 

\else

\fi

\title{PERL: Pivot-based Domain Adaptation for Pre-trained  Deep Contextualized Embedding Models}

\author{Eyal Ben-David \thanks{* Both authors equally contributed to this work.}\\
\And Carmel Rabinovitz \footnotemark[1] \\
 Technion, Israel Institute of Technology\\
{\tt \{eyalbd12@campus.$|$carmelrab@campus.$|$roiri@\}technion.ac.il}
\And Roi Reichart\\
}

\date{}
\begin{document}

\maketitle

\begin{abstract}
    Pivot-based neural representation models have lead to significant progress in domain adaptation for NLP. However, previous works that follow this approach utilize only labeled data from the source domain and unlabeled data from the source and target domains, but neglect to incorporate massive unlabeled corpora that are not necessarily drawn from these domains. To alleviate this, we propose \textit{PERL}: A representation learning model that extends contextualized word embedding  models such as BERT \cite{devlin2018bert} with pivot-based fine-tuning. PERL  outperforms strong baselines across 22 sentiment classification domain adaptation setups, improves  in-domain model performance, yields effective reduced-size models and increases model stability.\footnote{Our code is at \url{https://github.com/eyalbd2/PERL}.}
    \footnote{This paper was accepted to TACL in June 2020}
\end{abstract}

\section{Introduction}
\label{sec:intro}

Natural Language Processing (NLP) algorithms are constantly improving, gradually approaching human level performance \cite{Dozat:17,Edunov:18,Radford:18}. However, those algorithms often depend on the availability of large amounts of manually annotated data from the domain where the task is performed. Unfortunately, collecting such annotated data is often costly and laborious, which substantially limits the applicability of NLP technology.


Domain Adaptation (DA), training an algorithm on annotated data from a source domain so that it can be effectively applied to other target domains, is one of the ways to solve the above bottleneck. Indeed, over the years substantial efforts have been devoted to the DA challenge \citep{roark2003supervised, daume2006domain, ben2010theory, jiang2007instance, mcclosky2010automatic, rush2012improved, schnabel2014flors}. Our focus in this paper is on unsupervised DA, the setup we consider most realistic. In this setup labeled data is available only from the source domain while unlabeled data is available from both the source and the target domains.
 
While various approaches for DA have been proposed (\S \ref{sec:previous}), with the prominence of deep neural network (DNN) modeling, attention has been recently focused on representation learning approaches.
Within representation learning for unsupervised DA, two approaches have been shown particularly useful. In one line of work, DNN-based methods which employ compress-based noise reduction to learn cross-domain features have been developed \cite{glorot2011domain, chen2012marginalized}. In another line of work, methods based on the distinction between pivot and non-pivot features \cite{blitzer2006domain, blitzer2007biographies} learn a joint feature representation for the source and the target domains. Later on, \newcite{ziser2016neural,ziser2018pivot}, and \newcite{li2018hierarchical} married the two approaches and achieved substantial improvements on a variety of DA setups.

Despite their success, pivot-based DNN models still only utilize labeled
data from the source domain and unlabeled data from both the source and the target domains, but neglect to incorporate massive unlabeled corpora that are not necessarily drawn from these domains. With the recent game-changing success of contextualized word embedding models trained on such massive corpora \cite{devlin2018bert, peters2018deep}, it is natural to ask whether information from such corpora can enhance these DA methods, particularly that background knowledge from non-contextualized embeddings has shown useful for DA \citep{DBLP:conf/acl/PlankM13, DBLP:conf/acl/NguyenPG15}.

In this paper we hence propose an unsupervised DA approach that
extends leading approaches based on DNNs and pivot-based ideas, so that they can incorporate information encoded in massive corpora (\S \ref{sec:DA with PERL}). Our model, named \textit{PERL}: \textit{Pivot-based Encoder Representation of Language}, builds on massively pre-trained contextualized word embedding models such as BERT \cite{devlin2018bert}. To adjust the representations learned by these models so that they close the gap between the source and target domains, we fine-tune their parameters using a pivot-based variant of the Masked Language Modeling (MLM) objective, optimized on unlabeled data from both the source and the target domains. We further present R-PERL (regularized PERL) which
facilitates parameter sharing for pivots with similar meaning.

We perform extensive experimentation in various unsupervised DA setups of the task of binary sentiment classification (\S \ref{sec:experiments}, \ref{sec:results}). First, for compatibility with previous work, we experiment with the legacy product review domains of \newcite{blitzer2007biographies} (12 setups). We then experiment with more challenging setups, adapting between the above domains and the airline review domain \cite{Nguyen2015airline} used in \newcite{ziser2018pivot} (4 setups), as well as the IMDB movie review domain \cite{maas2011learning} (6 setups). We compare PERL to the best performing pivot-based methods \cite{ziser2018pivot,li2018hierarchical} and to DA approaches that fine-tune a massively pre-trained BERT model by optimizing its standard MLM objective using target-domain unlabeled data \cite{lee2019biobert,han2019unsupervised}.
PERL and R-PERL substantially outperform these baselines, emphasizing the additive effect of massive pre-training and pivot-based fine-tuning. 
%

As an additional contribution, we show that pivot-based learning is effective beyond improving domain adaptation accuracy. Particularly, we show that an in-domain variant of PERL substantially improves the in-domain performance of a BERT-based sentiment classifier, for varying training set sizes (from 100 to 20K labeled examples). We also show that PERL facilitates the generation of effective reduced-size DA models. Finally, we perform an extensive ablation study (\S \ref{sec:ablation}) that uncovers PERL's crucial design choices and demonstrates the stability of PERL to hyper-parameter selection compared to other DA methods. 


\section{Background and Previous Work}
\label{sec:previous}

There are several approaches to DA, including instance re-weighting \citep{sugiyama2008direct, huang2007correcting, mansour2009domain}, sub-sampling from the participating domains \citep{chen2011automatic} and DA through representation learning, where a joint representation is learned based on texts from the source and target domains \citep{blitzer2007biographies,xue2008topic,ziser2016neural,ziser2018pivot}.  We first describe the unsupervised DA pipeline, continue with representation learning methods for DA with a focus on pivot-based methods, and, finally, describe contextualized  embedding models. 

\paragraph{Unsupervised Domain Adaptation through Representation Learning}

As said in \S \ref{sec:intro}  our focus in this work is on unsupervised DA through representation learning. A common pipeline for this setup consists of two
steps: (A) Learning a representation model (often referred to as the encoder) using the source and target unlabeled data; and (B) Training a supervised classifier on the source domain labeled data. To facilitate domain adaptation, every text fed to the classifier in the second step is first represented by the pre-trained encoder. This is performed both when the classifier is trained in the source domain and when it is applied to new text from the target domain.

Exceptions to this pipeline are end-to-end
models that jointly learn to perform the cross-domain text representation and the classification task. This is achieved by training a unified objective on the source domain labeled data and the unlabeled data from both the source and the target. Among these models are domain adversarial networks \cite{Ganin:16}, which were strongly outperformed by \newcite{ziser2018pivot} to which we compare our methods, and the  hierarchical  attention  transfer  network (HATN, \cite{li2018hierarchical}), which is one of our baselines (see below).

Unsupervised DA through representation learning has followed two main avenues. The first avenue consists of works that aim to explicitly build a feature representation that bridges the gap between the domains. A seminal framework in this line is structural correspondence learning (SCL, \citep{blitzer2006domain, blitzer2007biographies}), that splits the feature space into pivot and non-pivot features. A large number of works have followed this idea (e.g. \cite{pan2010cross, gouws2012learning,bollegala2015unsupervised,yu2016learning, DBLP:conf/ijcai/LiZWWY17,li2018hierarchical, DBLP:journals/access/TuW19,  ziser2016neural,ziser2018pivot}) and we discuss it below. 

Works in the second avenue learn cross-domain representations by training autoencoders (AEs) on the unlabeled data from the source and target domains. This way they hope to get a more robust representation, which is hopefully better suited for DA.
Examples for such models include the stacked denoising AE (SDA, \citep{vincent2008extracting, glorot2011domain}, the marginalized SDA and its variants (MSDA, \citep{chen2012marginalized, yang2014fast, clinchant2016domain}) and variational AE based models  \citep{louizos2015variational}. 

Recently, \newcite{ziser2016neural,ziser2018pivot} and \newcite{li2018hierarchical} married these approaches and presented pivot-based approaches where the representation model is based on DNN encoders (AE, LSTM or hierarchical attention networks). Since their methods outperformed the above models, we aim to extend them to models that can also exploit massive out of (source and target) domain corpora. We next elaborate on pivot-based approaches.

\paragraph{Pivot-based Domain Adaptation}



Proposed by \citet{blitzer2006domain, blitzer2007biographies} through their SCL framework, the main idea of pivot-based DA is to divide the shared feature space of the source and the target domains to two complementary subsets: one of pivots and one of non-pivots. Pivot features are defined based on two criteria: (a) They are frequent in the unlabeled data of both domains; and (b) They are prominent for the classification task defined by the source domain labeled data. Non-pivot features are those features that do not meet at least one of the above criteria. While SCL is based on linear models, there have been some very successful recent efforts to extend this framework so that non-linear encoders (DNNs) are employed. Here we focus on the latter line of work, which produces much better results, and do not elaborate on SCL any further.

%


\newcite{ziser2018pivot} have presented the Pivot Based Language Model (PBLM), which incorporates pre-training and pivot-based learning. PBLM is a variant of an LSTM-based language model, but instead of predicting at each point the most likely next input word, it predicts the next input unigram or bigram if one of these is a pivot (if both are, it predicts the bigram), and NONE otherwise. In the unsupervised DA pipeline PBLM is trained on the source and target unlabeled data. Then, when the task classifier is trained and applied to the target domain, PBLM is employed as a contextualized word embedding layer. Notice that PBLM is not pre-trained on massive out of (source and target) domain corpora, and its single-layer, unidirectional LSTM architecture is probably not ideal for knowledge encoding  from such corpora.

Another work in this line is HATN \cite{li2018hierarchical}. This model automatically learns the pivot/non-pivot distinction, rather than following the SCL definition as \newcite{ziser2016neural,ziser2018pivot} did. 
HATN consists of two hierarchical attention networks, P-net and NP-net. First, it trains the P-net on the source labeled data. Then, it decodes the most prominent tokens of P-net (i.e. tokens which received the highest attention values), and considers them as its pivots. Finally, it simultaneously trains the P-net and the NP-net on both the labeled and the unlabeled data, such that P-net is adversarially trained to predict the domain of the input example \cite{Ganin:16} and NP-net is trained to predict its pivots, and the hidden representations from both networks serve for the task label (sentiment) prediction.
%
%


Both HATN and PBLM strongly outperform a large variety of previous DA models on various cross-domain sentiment classification setups. Hence, they are our major baselines in this work. Like PBLM, we employ the same definition of the pivot and non-pivot subsets as in \newcite{blitzer2007biographies}. Like HATN, we also employ an attention-based DNN. Unlike both models, we design our model so that it incorporates pivot-based learning with pre-training on massive out of (source and target) domain corpora. We next discuss this pre-training process, which is also known as training models for contextualized word embeddings.

\paragraph{Contextualized Word Embedding Models}
\label{sec:CWE}


Contextualized word embedding (CWE) models are trained on massive corpora \citep{peters2018deep, radford2019language}. They typically employ a language modeling objective or a closely related variant \citep{peters2018deep, ziser2018pivot, devlin2018bert, yang2019xlnet}, although in some recent papers the model is trained on a mixture of basic NLP tasks \citep{zhang2019ernie,rotmandeep}. The contribution of such models to the state-of-the-art in a variety of NLP tasks is already well-established. 

CWE models typically follow three steps: \textbf{(1)} Pre-training: Where a DNN (referred to as the encoder of the model) is first trained on massive unlabeled corpora which represent a broad domain (such as English Wikipedia); \textbf{(2)} Fine-tuning: An optional step, where the encoder is refined on unlabeled text of interest. As noted above, \newcite{lee2019biobert} and \newcite{han2019unsupervised} tuned BERT on unlabeled target domain data to facilitate domain adaptation; and \textbf{(3)} Supervised task training: Where task specific layers are trained on labeled data for a downstream task of interest. 

PERL employs a pre-trained encoder, BERT  in this paper. BERT's architecture is based on  multi-head attention layers, trained with a two-component objective: (a) MLM and (b) Is-next-sentence prediction (NSP). For Step 2, PERL modifies only the MLM objective and it can hence be implemented within any CWE framework that employs this objective  \cite{liu2019roberta,lan2019albert,yang2019xlnet}. 



MLM is a modified language modeling objective, adjusted to self-attention models. 
%
%
When building the pre-training task, all input tokens  have the same probability to be masked.\footnote {We use the \emph{huggingface} BERT code \cite{Wolf2019HuggingFacesTS}: \url{https://github.com/huggingface/transformers}, where the masking probability is 0.15.} After the masking process, the model has to predict a distribution over the vocabulary for each masked token given the non-masked tokens. The input text may have more than one masked token, and when predicting one masked token information from the other masked tokens is not utilized.





In the next section we describe our PERL domain adaptation model. The novel component of this model is a pivot-based MLM objective, optimized at the fine-tuning step (Step 2) of the CWE pipeline, using source and target unlabeled data.

\begin{figure*}[ht]
 
\begin{subfigure}{0.33\textwidth}
\includegraphics[scale=0.3]{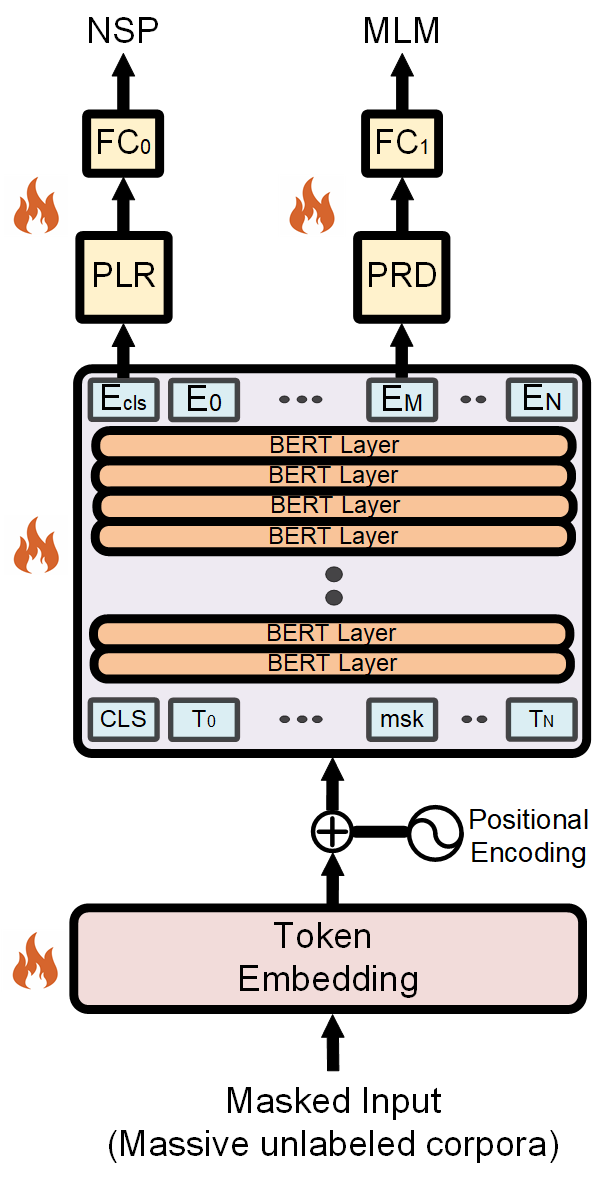}
\centering
\caption{Pre-training model}
\centering
\label{fig:BERT}
\end{subfigure}
\begin{subfigure}{0.33\textwidth}
\includegraphics[scale=0.3]{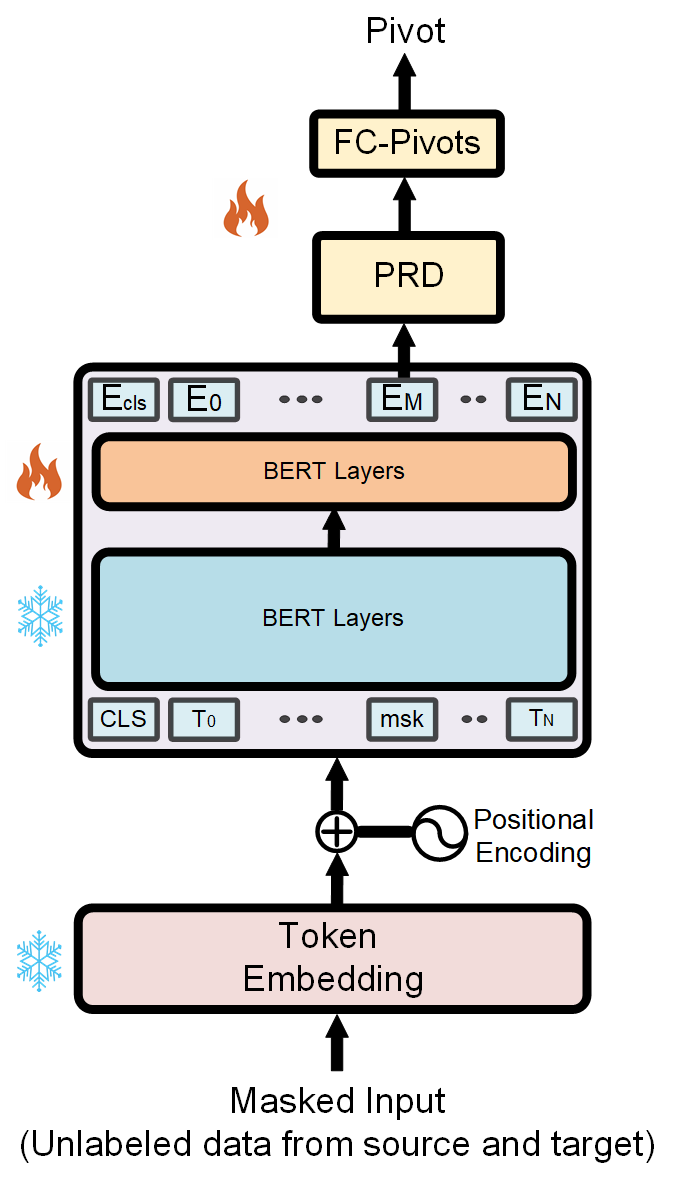}
\centering
\caption{Fine-tuning model}
\centering
\label{fig:PERL-fine-tune}
\end{subfigure}
\begin{subfigure}{0.33\textwidth}
\includegraphics[scale=0.3]{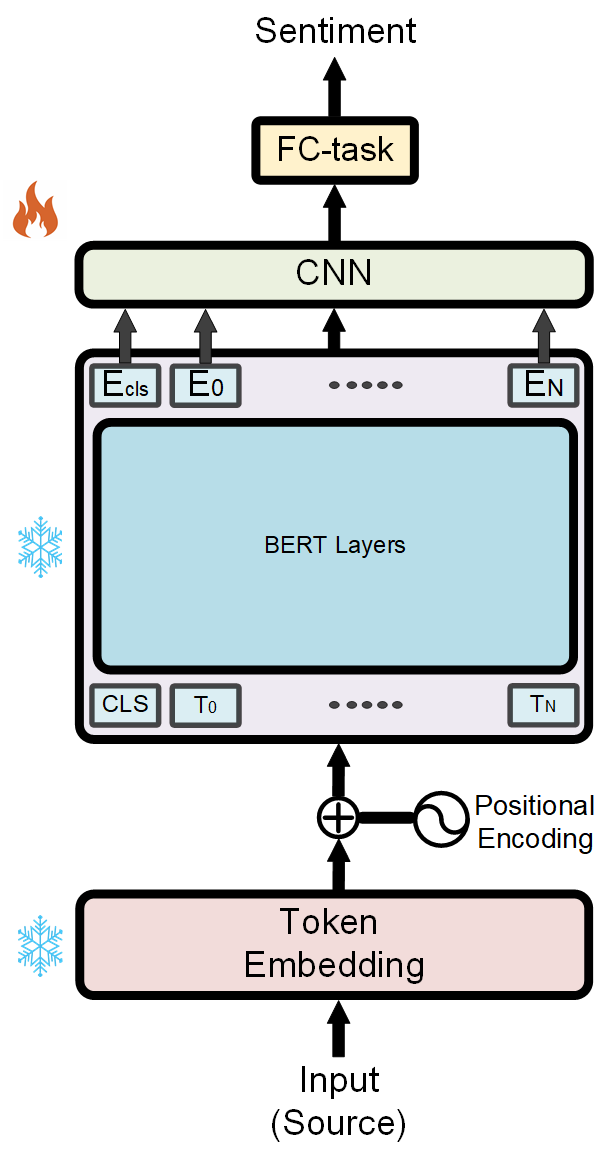}
\centering
\caption{Task specific model}
\centering
\label{fig:task_specific}
\end{subfigure}
\caption{Illustrations of the three PERL steps. PRD and PLR stand for the BERT prediction head and pooler head respectively, FC is a fully connected layer, and msk stands for masked tokens embeddings (embeddings of tokens that were masked). NSP and MLM are the next sentence prediction and masked language model objectives. For the definitions of the PRD and PRL layers as well as the NSP objective, see \newcite{devlin2018bert}. We mark frozen layers (layers whose parameters are kept fixed) and non-frozen layers with snow-flake and fire symbols, respectively. 
The token embedding and BERT layers values at the end of each step initialize the corresponding layers of the next step model. The BERT box of the fine tuning step is described in more details in Figure~\ref{fig:perl-pivot-task}.}
\centering
\label{fig:all_stages}
\end{figure*}

\section{Domain adaptation with PERL}
\label{sec:DA with PERL}

PERL employs pivot features in order to learn a representation that bridges the gap between two domains. Contrary to previous pivot-based DA representation models, it exploits unlabeled data from the source and target domains, and also from massive out of source and target domain corpora. 

PERL consists of three steps that correspond to the three steps of CWE models, as described in \S~\ref{sec:CWE}: \textit{(1) Pre-training (Figure~\ref{fig:BERT})}: in which it employs a pre-trained CWE model (encoder, BERT in this work) that was trained on massive corpora; \textit{(2) Fine-tuning (Figure~\ref{fig:PERL-fine-tune})}: where it refines some of the pre-trained encoder weights, based on a pivot-based objective that is optimized on unlabeled data from the source and target domains; and \textit{(3) Supervised task training (Figure~\ref{fig:task_specific})}: where task specific layers are trained on source domain labeled data for the downstream task of interest.

Our pivot selection method is identical to that of \newcite{blitzer2007biographies} and \newcite{ziser2016neural,ziser2018pivot}. That is, the pivots are selected independently of the above three steps protocol. 

We further present a variant of PERL, denoted with R-PERL, where the non-contextualized embedding matrix of the BERT model trained at Step (1) is employed in order to regularize PERL during its fine-tuning stage (Step 2). We elaborate on this model towards the end of this section. We next provide a detailed description. 

\paragraph{Pivot Selection}

Being a pivot-based language representation model, PERL is based on high quality pivot extraction. Since the representation learning is based on a masked language modeling task, the feature set we address consists of the unigrams and bigrams of the vocabulary. We base the division of this feature set into pivots and non-pivots on unlabeled data from the source and target domains. Pivot features are: (a) Frequent in the unlabeled data from the source and target domains; and (b) Among those frequent features, pivot features are the ones whose mutual information with the task label according to source domain labeled data crosses a pre-defined threshold. Features that do not meet the above two criteria form the non-pivot feature subset. 


\paragraph{PERL pre-training (Step 1, Figure~\ref{fig:BERT})}

In order to inject prior language knowledge to our model, we first initialize the PERL encoder with a powerful pre-trained CWE model. As noted above, our rationale is that the general language knowledge encoded in these models, which is not specific to the source or target domains,  should be useful for DA just as it has shown useful for in-domain learning. In this work we employ BERT, although any other CWE model that employs the MLM objective for pre-training (Step 1) and fine-tuning (Step 2), could have been used. 

\com{
\begin{algorithm}[!t]
\caption{\footnotesize PERL Training Process}\label{alg:PERL training}
{\footnotesize
\textbf{Input:}
Source labeled data $\mathbf{L_{s}}$, Source unlabeled data $\mathbf{U_{s}}$, Target unlabeled data $\mathbf{U_{t}}$, General domain raw text $\mathbf{U_{g}}$\\   

$\textbf{Algorithm:}$
\begin{enumerate}
    \item Pre-train - Train the encoder on $\mathbf{U_{g}}$ (\S~\ref{sec:CWE}; \S~\ref{sec:PERL_Init}), we merely initialize with pre-trained BERT.
    \item Fine-tune - refine the encoder on  $\mathbf{U_{s}}$ and $\mathbf{U_{t}}$, using pivot based MLM objective  (\S~\ref{sec:PERL_fintuning}).
    \item Task specific, train task specific layers on $\mathbf{L_{s}}$  (\S~\ref{sec:PERL_task_specific}).
\end{enumerate}
}
\end{algorithm}
}

\paragraph{PERL fine-tuning (Step 2, Figure~\ref{fig:PERL-fine-tune})}

This step is the core novelty of PERL. Our goal is to refine the initialized encoder on unlabeled data from the source and the target domains, using the distinction between pivot and non-pivot features. 

\begin{figure}[t!]
\includegraphics[scale=0.4]{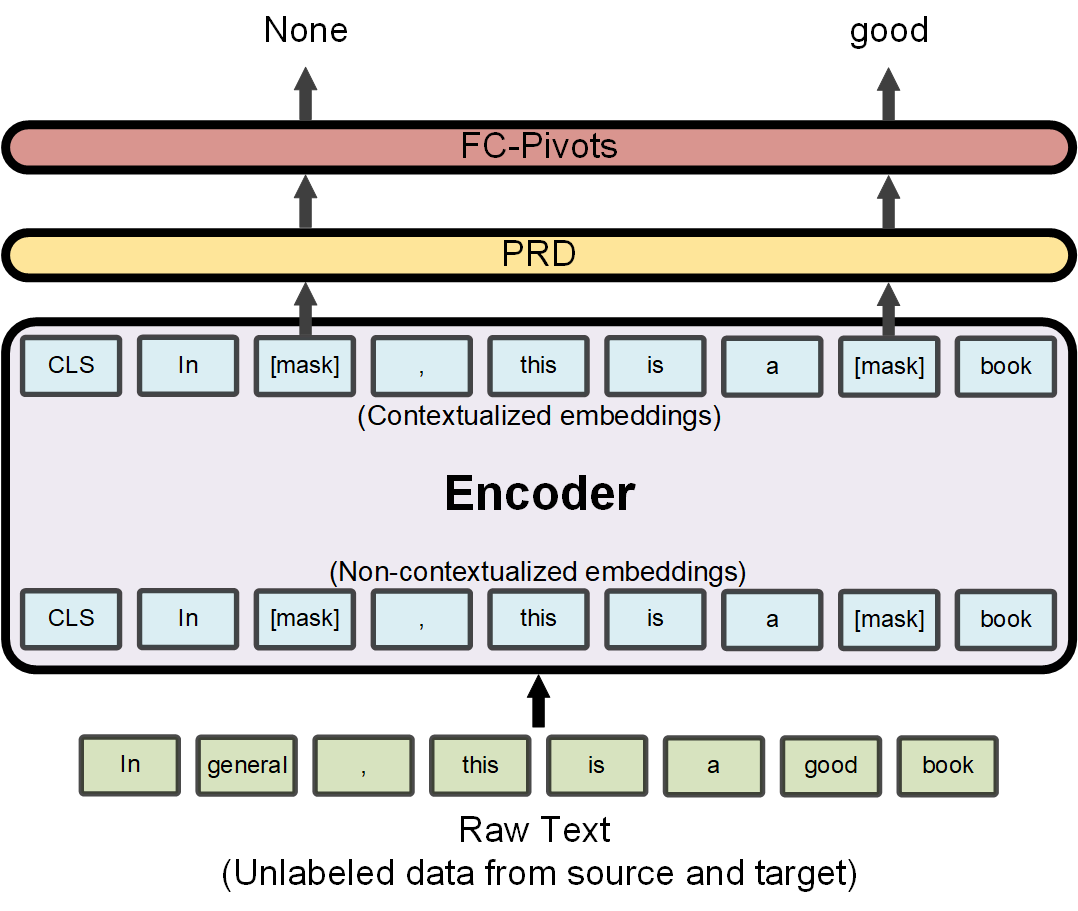}
\centering
\caption{The PERL pivot-based fine-tuning task (Step 2). In this example two tokens are masked, \textit{general} and \textit{good}, only the latter is a pivot. The architecture is identical to that of BERT but the MLM task and the masking process are different, taking into account the pivot/non-pivot distinction.}
\centering
\label{fig:perl-pivot-task}
\end{figure}

For this aim we fine-tune the parameters of the pre-trained BERT using its MLM objective, but we choose the masked words so that the model learns to map non-pivot to pivot features. Recall, that when building the MLM training task, each training example consists of an input text in which some of the words are masked, and the task of the model is to predict the identity of each of the masked words given the rest of the (non-masked) input text. While in standard MLM training all input tokens have the same probability to be masked, in the PERL fine-tuning step we change both the masking  probability and the prediction task so that the desired non-pivot to pivot mapping is learned. We next describe these two changes, see also a detailed graphical illustration  in Figure~\ref{fig:perl-pivot-task}.

\textit {1. Prediction task.} While in standard MLM the task is to predict a token out of the entire vocabulary, here we define a pivot-base prediction task. Particularly, the model should predict whether the masked token is a pivot feature or not, and if it is then it has to identify the pivot. That is, this is a multi-class classification task where the number of classes is equal to the number of pivots plus 1 (for the non-pivot prediction). 

Put it more formally, the modified pivot-based MLM objective is:
\begin{equation} \nonumber
\label{eq:1}
p(y_i = j) = \dfrac{e^{f(h_i) \cdot W_j}} {\sum_{k=1}^{|P|}{e^{f(h_i) \cdot W_k}} + e^{f(h_i) \cdot W_{none}}}
\end{equation}

\noindent where $y_i$ is a masked unigram or bigram at position $i$, $P$ is the set of pivot features (token unigrams and bigrams), $h_i$ is the encoder representation for the $i$-th token, $W$ (the FC-Pivots layer of Figure~\ref{fig:PERL-fine-tune} and Figure~\ref{fig:perl-pivot-task}) is the pivot predictor matrix that maps from the latent space to the pivot set space ($W_a$ is the $a$-th row of $W$), and $f$ is a non-linear function composed of a \emph{dense} layer, a \emph{gelu} activation layer and \emph{LayerNorm} (the PRD layer of Figure~\ref{fig:PERL-fine-tune} and Figure~\ref{fig:perl-pivot-task}). 


\textit {2. Masking process.} Instead of masking each input token (unigram) with the same probability, we perform the following masking process. For each input token (unigram) we first check whether it forms a bigram pivot together with the next token, and if so we mask this bigram with a probability of $\alpha$. If the answer is negative, we check if the token at hand is a unigram pivot and if so we again mask it with a probability of $\alpha$. Finally, if the token is not a pivot we mask it with a  probability of $\beta$. Our hyper-parameter tuning process revealed that the values of $\alpha = 0.5$ and $\beta = 0.1$ provide strong results across our various experimental setups (see more on this in \S \ref{sec:ablation}). This way PERL gives a higher probability to pivot masking, and by doing so the encoder parameters are fine-tuned so that they can predict (mostly) pivot features based (mostly) on non-pivot input. 

Designing the fine-tuning task this way yields two advantages. First, the model should shape its parameters so that most of the information about the input pivots is preserved, while most of the information preserved about the non-pivots is what needed in order to predict the existence of the pivots. This way the model keeps mostly the information about unigrams and bigrams that are shared among the two domains and are significant for the supervised task, thus hopefully increasing its cross-domain generalization capacity. 

Second, standard MLM, which has recently been used for  fine-tuning in domain adaptation \cite{lee2019biobert, han2019unsupervised}, performs a multi-class classification task with 30K tokens,\footnote{The BERT implementation we use keeps a fixed 30K word vocabulary, derived from its pre-training process.} which requires $\sim 23$M parameters as in the FC1 layer of Figure~\ref{fig:BERT}. By focusing PERL on pivot prediction, we can use only a factor of $\frac{|P|+1}{30K}$  of the FC layer parameters, as we do in the FC-pivots layer (Figure~\ref{fig:PERL-fine-tune}, where $|P|$ is the number of pivots, in our experiments $|P| \in{[100,500]} $). 



\paragraph{Supervised task training (Step 3,  Figure~\ref{fig:task_specific})}

To adjust PERL for a downstream task, we place a classification network on top of its encoder. While training on labeled data from the source domain and testing on the target domain, each input text is first represented by the encoder and is then fed to the classification network. Since our focus in this work is on the representation learning, the classification network is kept simple, consisting of one convolution layer followed by an average pooling layer and a linear layer. When training for the downstream task, the encoder weights are frozen. 

\paragraph{R-PERL}

A potential limitation of PERL is that it ignores the semantics of its pivots. While the negative pivots \textit{sad} and \textit{unhappy} encode similar information with respect to the sentiment classification task, PERL considers them as two different output classes. To alleviate this, we propose the regularized PERL (R-PERL) model where pivot-similarity information is taken into account. 

To achieve this we construct the FC-pivots matrix of R-PERL (Figures~\ref{fig:PERL-fine-tune} and ~\ref{fig:perl-pivot-task}) based on the Token Embedding matrix learned by BERT in its pre-training stage (Figure~\ref{fig:BERT}). Particularly, we fix the unigram pivot rows of the FC-pivots matrix to the corresponding rows in BERT's Token Embedding matrix, and the bigram pivot rows to the mean of the Token Embedding rows that correspond to the unigrams that form this bigram. The FC-pivots matrix of R-PERL is kept fixed during  fine-tuning. 

Our assumptions are that: (1) Pivots with similar meaning, such as \textit{sad} and \textit{unhappy} have similar representations in the Token Embedding matrix learned at the pre-training stage (Step 1); and (2) There is a positive correlation between the appearance of such pivots (i.e. they tend to appear, or not appear, together; see \cite{ziser2016neural} for similar considerations). In its fine-tuning step, R-PERL is hence biased to learn similar representations to such pivots in order to capture the positive correlation between them. This follows from the fact that  pivot probability is computed by taking the dot product of its representation with its corresponding row in the FC-pivots matrix.

%
%

\com{
\begin{figure}[ht]
\includegraphics[scale=0.4]{figures/LM_prediction_head.png}
\centering
\caption{Architecture of a prediction-head (PRD). The PRD is part of the fine-tuning stage and the pre-training stage. This architecture is not used in task specific stage.}
\centering
\label{fig:prediction_head}
\end{figure}
}

\section{Experiments}
\label{sec:experiments}

\paragraph{Tasks and Domains}

Following a large body of prior DA work, we focus on the task of binary sentiment classification. For compatibility with previous literature, we first experiment with the four legacy product review domains of \newcite{blitzer2007biographies}: Books (B), DVDs (D), Electronic items (E) and Kitchen appliances (K) with a total of 12 cross-domain setups. Each domain has 2000 labeled reviews, 1000 positive and 1000 negative, and unlabeled reviews as follows: B: 6,000, D: 34,741, E: 13,153 and K: 16,785. 

We next experiment in a more challenging setup, considering an airline review dataset (A) \cite{Nguyen2015airline, ziser2018pivot}. This setup is challenging both due to the differences between the product and service domains, and because the prior probability of observing a  positive review at the A domain is much lower than the same probability in the product domains.\footnote{This analysis, performed by \newcite{ziser2018pivot}, is based on the gold labels of the unlabeled data.} 
For the A domain, following \newcite{ziser2018pivot}, we randomly sampled 1000 positive and 1000 negative reviews for our labeled set, and 39396 reviews for our unlabeled set.  Due to the heavy computational demands of the experiments, we arbitrarily chose 3 product to airline and 3 airline to product setups. 

We further consider an additional modern domain: IMDB (I) \citep{maas2011learning},\footnote{The details of the IMDB dataset are available at: \url{http://www.andrew-maas.net/data/sentiment} .} which is commonly used in recent sentiment analysis work. This dataset consists of 50000 movie reviews from IMDB (25000 positive and 25000 negative), where there is a limitation on the number of reviews per movie. We  randomly  sampled 2000 labeled reviews, 1000 positive and 1000 negative, for our labeled set, and the remaining 48000 reviews form our unlabeled set.\footnote{We make sure that all reviews of the same movie appear either in the training set or in the test set.} As above, we arbitrarily chose 2 IMDB to product and 2 product to IMDB setups for our experiments.

Pivot-based representation learning has shown instrumental for DA. We hypothesize that it can also be beneficial for in-domain tasks, as it focuses the representation on the information encoded in prominent unigrams and bigrams.
To test this hypothesis we experiment in an in-domain setup, with the IMDB movie review dataset. We follow the same experimental setup as in the domain adaptation case, except that only IMDB unlabeled data is used for fine-tuning, and the frequency criterion in pivot selection is defined with respect to this dataset.

We randomly sampled 25000 training and 25000 test examples, keeping the two sets balanced, and additional 50000 reviews formed an unlabeled balanced set.\footnote{These reviews are also part of the IMDB dataset.} We consider 6 setups, differing in their training set size: 100, 500, 1K, 2K, 10K and 20K randomly sampled examples.

\paragraph{Baselines}

We compare our PERL and R-PERL models to the following baselines: (a+b) PBLM-CNN and PBLM-LSTM \cite{ziser2018pivot}, differing only in their classification layer (CNN vs. LSTM);\footnote{ \url{https://github.com/yftah89/PBLM-Domain-Adaptation}} (c) HATN \cite{li2018hierarchical};\footnote{ \url{https://github.com/hsqmlzno1/HATN}}  (d) BERT; and (e) Fine-tuned BERT (following \cite{lee2019biobert,han2019unsupervised}): This model is identical to PERL, except that the fine-tuning stage is performed with a standard MLM instead of our pivot-based MLM. BERT, Fine-tuned BERT, PBLM-CNN, PERL and R-PERL all use the same CNN-based sentiment classifier, while HATN jointly learns the feature representation and performs sentiment classification.


\begin{table*}[t]
\begin{adjustbox}{width=\textwidth}
\begin{tabular}{ c | c | c | c | c | c | c | c | c | c | c | c | c || c }
& {D $\rightarrow$ K} & {D $\rightarrow$ B} & {E $\rightarrow$ D} & {B $\rightarrow$ D} & {B $\rightarrow$ E} & {B $\rightarrow$ K} & {E $\rightarrow$ B} & {E $\rightarrow$ K} &  {D $\rightarrow$ E} & {K $\rightarrow$ D} &  {K $\rightarrow$ E} &  {K $\rightarrow$ B} & {ALL} \\
    \hline
     BERT & 82.5 & 81.0 & 76.8 & 80.6 & 78.8 & 82.0 & 78.2 & 85.1 & 76.5 & 77.7 & 84.7 & 78.5 & 80.2 \\ 
     Fine-tuned BERT & 86.9 & 84.1 & 81.7 & 84.4 & 84.2 & 86.7 & 80.2 & 89.2 & 82.0 & 79.8 & 88.6 & 81.5 & 84.1 \\ 
    \hline
    PBLM-Max & 83.3 & 82.5 & 77.6 & 84.2 & 77.6 & 82.5 & 71.4 & 87.8 & 80.4 & 79.8 & 87.1 & 74.2 & 80.7 \\
     HATN & 85.4 & 83.5 & 78.8 & 82.2 & 78.0 & 81.2 & 80.0 & 87.4 & 83.2 & 81.0 & 85.9 & 81.2 & 82.3\\ 
    \hline 
    \hline
     PERL & 89.9 & 85.0 & \textbf{85.0} & 86.5 & 87.0 & 89.9 & \textbf{84.3} & 90.6 & 87.1 & 84.6 & 90.7 & 81.9 & 86.9 \\ 
     R-PERL & \textbf{90.4} & \textbf{85.6} & 84.8 & \textbf{87.8} & \textbf{87.2} & \textbf{90.2} & 83.9 & \textbf{91.2} & \textbf{89.3} & \textbf{85.6} & \textbf{91.2} & \textbf{83.0} & \textbf{87.5} \\ 
\end{tabular}
\end{adjustbox}
\end{table*}

\begin{table*}[t]
\centering
\begin{adjustbox}{width=\textwidth}
\begin{tabular}{ c | c | c | c | c || c || c | c | c | c | c | c || c }
& {I $\rightarrow$ E} & {I $\rightarrow$ K} & {E $\rightarrow$ I} & {K $\rightarrow$ I} & {ALL} & {A $\rightarrow$ B} & {A $\rightarrow$ K} & {A $\rightarrow$ E} & {B $\rightarrow$ A} &  {K $\rightarrow$ A} & {E $\rightarrow$ A} & {ALL}  \\
    \hline
     BERT & 75.4 & 78.8 & 72.2 & 70.6 & 74.2 & 70.9 & 78.8 & 77.1 & 72.1 & 74.0 & 81.0 & 75.6 \\ 
     Fine-tuned BERT & 81.5 & 78.0 & 77.6 & 78.7 & 78.9 & 72.9 & 81.9 & 83.0 & 79.5 & 76.3 & 82.8 & 79.4 \\
    \hline
    PBLM-Max & 70.1 & 69.8 & 67.0 & 69.0 & 69.0 & 70.6 & 82.6 & 81.1 & 83.8 & \textbf{87.4} & \textbf{87.7} & 80.5 \\ 
     HATN & 74.0 & 74.4 & 74.8 & 78.9 & 75.5 & 58.7 & 68.8 & 64.1 & 77.6 & 78.5 & 83.0 & 71.8 \\ 
    \hline
    \hline
     PERL & 87.1 & \textbf{86.3} & 82.0 & 82.2 & 84.4 & 77.1 & 84.2 & 84.6 & 82.1 & 83.9 & 85.3 & 82.9 \\
     R-PERL & \textbf{87.9} & 86.0 & \textbf{82.5} & \textbf{82.5} & \textbf{84.7} & \textbf{78.4} & \textbf{85.9} & \textbf{85.9} & \textbf{84.0} & 85.1 & 85.9 & \textbf{84.2} \\

\end{tabular}
\end{adjustbox}
\caption{Domain adaptation results. The top table is for the legacy product review domains of \newcite{blitzer2007biographies} (denoted as the $P \Leftrightarrow P$ setups in the text). The bottom table involves selected legacy domains as well as the IMDB movie review domain (left; denoted as $P \Leftrightarrow I$) or the airline review domain (right; denoted as $P \Leftrightarrow A$). The All columns present averaged results across the setups to their left.}
\label{tab:Classification accuracy on the Amazon review dataset.}
\end{table*}


\begin{table}
\centering
\begin{adjustbox}{width=0.38\textwidth}
\begin{tabular}{c | c | c | c | c }
Num &   & Fine-tuned &  \\
Sentences & BERT  & BERT & PERL & R-PERL \\
    \hline
     100 & 67.9 & 76.4 & 81.6 &\textbf{83.9} \\ 
     500 & 73.9 &  83.3 & 84.3 & \textbf{84.6} \\
     1K & 75.3 &  83.9 & 84.6 & \textbf{84.9} \\
     2K & 77.9 &  83.6 &  \textbf{85.3} & \textbf{85.3} \\
     10K & 80.9 &  86.9 & 87.1 & \textbf{87.5} \\
     20K & 81.7 &  86.0 & 87.8 & \textbf{88.1} \\
\end{tabular}
\end{adjustbox}
\caption{In domain results on the IMDB movie review domain with increasing training set size.}
\label{tab:In Domain Classification accuracy on the IMDB dataset.}
\end{table}

\paragraph{Cross-validation}

We employ a five fold cross-validation protocol, where in every fold 80\% of the source domain examples are randomly selected for training data, and 20\% for development data (both sets are kept balanced). For each model we report the average results across the five folds. In each fold we tune the hyper-parameters so that to minimize the cross-entropy development data loss. 

\paragraph{Hyper-parameter Tuning}

For all models we use the WordPiece word embeddings \cite{DBLP:journals/corr/WuSCLNMKCGMKSJL16} with a vocabulary size of 30k, and the same optimizer (with the same hyper-parameters) as in their original paper. For all pivot-based methods we consider the unigrams and bigrams that appear at least 20 times both in the unlabeled data of the source domain and in the unlabeled data of the target domain as candidates for pivots,\footnote{In the in-domain experiments we consider the IMDB unlabeled data.} and from these we select the $|P|$ candidates with the highest mutual information with the task source domain label ($|P| = \{100, 200, \ldots, 500\}$). The exception is HATN that automatically selects its pivots, which are limited to unigrams.

We next describe the hyper-parameters of each of the models. Due to our extensive experimentation (22 DA and 6 in-domain setups, 5-fold cross-validation), we limit our search space, especially for the heavier components of the models.

\textit{R-PERL, PERL, BERT and Fine-tuned BERT}  For the encoder, we use the BERT-base uncased architecture with the same hyper-parameters as in \newcite{devlin2018bert}, tuning for PERL, R-PERL and Fine-tuned BERT the number of fine-tuning epochs (out of: 20, 40, 60) and the number of unfrozen BERT layer during the fine-tuning process (1, 2, 3, 5, 8, 12). For PERL and R-PERL we tune the number of pivots (100, 200, 300, 400, 500) as well as $\alpha$ and $\beta$ (0.1, 0.3, 0.5, 0.8). The supervised task classifier is a basic CNN architecture, which enables us to search over the number of filters (out of: 16, 32, 64), the filter size (7, 9, 11) and the training batch size (32, 64).   


\textit{PBLM-LSTM and PBLM-CNN} For PBLM we tune the input word embedding size (32, 64, 128, 256), the number of pivots (100, 200, 300, 400, 500) and the hidden dimension (128, 256, 512). For the LSTM classification layer of PBLM-LSTM we consider the same hidden dimension and input word embedding size as for the PBLM encoder. For the CNN classification layer of PBLM-CNN, following \newcite{ziser2018pivot} we use 250 filters and a kernel size of 3. In each setup we choose the PBLM model (PBLM-LSTM or PBLM-CNN) that yields better test set accuracy 
and report its result, under PBLM-Max.



\textit{HATN} The hyper-parameters of \newcite{li2018hierarchical} were tuned on a larger training set than ours, and  they hence yield sub-optimal performance in our setup. We tune the training batch size (20, 50 300), the hidden layer size (20, 100, 300) and the word embedding size (50, 100, 300).

\section{Results}
\label{sec:results}




\paragraph{Overall results}

Table \ref{tab:Classification accuracy on the Amazon review dataset.} presents domain adaptation results, and is divided to 2 sub-tables. The top table reports results on the 12 setups derived from the 4 legacy product review domains of \newcite{blitzer2007biographies} (denoted with $P \Leftrightarrow P$). The bottom table reports results for 10 setups involving product review domains and the IMDB movie review domain (left side; denoted $P \Leftrightarrow I$) or the airline review domain (right side; denoted $P \Leftrightarrow A$). 
Table \ref{tab:In Domain Classification accuracy on the IMDB dataset.} presents in-domain results on the IMDB domain, for various training set sizes. 



\paragraph{Domain Adaptation}

As presented in Table \ref{tab:Classification accuracy on the Amazon review dataset.}, PERL models are superior in 20 out of 22 DA setups, with R-PERL performing best in 17 out of 22 setups. 
In the $P \Leftrightarrow P$ setups, their averaged performance (top table, All column) are 87.5\% and 86.9\% (for R-PERL and PERL, respectively) compared to 82.3\% of HATN and 80.7\% of PBLM-Max. Importantly, in the more challenging setups, the performance of one of these baselines substantially degrade. Particularly, the averaged R-PERL and PERL performance in the $P \Leftrightarrow I$ setups are 84.7\% and 84.4\%, respectively (bottom table, left All column), compared to 75.5\% of HATN and 69.0\% of PBLM-Max. In the $P \Leftrightarrow A$ setups the averaged R-PERL and PERL performances are 84.2\% and 82.9\%, respectively (bottom table, right All column), compared to 80.5\% of PBLM-Max and only 71.8\% of HATN. 

The performance of BERT and Fine-tuned BERT also degrade on the challenging setups: From an average of 80.2\% (BERT) and 84.1\% (Fine-tuned BERT) in $P \Leftrightarrow P$ setups, to 74.2\% and 78.9\% respectively in $P \Leftrightarrow I$ setups, and to 75.6\% and 79.4\% respectively in $P \Leftrightarrow A$ setups. R-PERL and PERL, in contrast, remain stable across setups, with an averaged accuracy of 84.2-87.5\% (R-PERL) and 82.9-86.8\% (PERL).

The IMDB and airline domains differ from the product domains in their topic (movies (IMDB) and services (airline) vs. products). Moreover, the unlabeled data from the airline domain contains an increased fraction of negative reviews (see \S \ref{sec:experiments}). Finally, the IMDB and airline reviews are also more recent. The success of PERL in the $P \Leftrightarrow I$ and $P \Leftrightarrow A$ setups is of particular importance, as it indicates the potential of our algorithm to adapt supervised NLP algorithms to domains that substantially differ from their training domain.


Finally, our results clearly indicate the positive impact of a pivot-aware approach when fine-tuning BERT with unlabeled source and target data. Indeed, the averaged gaps between Fine-tuned BERT and BERT (3.9\% for $P \Leftrightarrow P$, 4.7\% for $P \Leftrightarrow I$ and 3.8\% for $P \Leftrightarrow A$) are much smaller than the corresponding gaps between R-PERL and BERT (7.3\% for $P \Leftrightarrow P$, 10.5\% for $P \Leftrightarrow I$ and 8.6\% for $P \Leftrightarrow A$). 



\paragraph{In-domain Results}

In this setup both the labeled and the unlabeled data, used for supervised task training (labeled data, Step 3), fine-tuning (unlabeled data, Step 2), and pivot selection (both datasets) come from the same domain (IMDB). As shown in Table \ref{tab:In Domain Classification accuracy on the IMDB dataset.}, PERL outperforms BERT and Fine-tuned BERT for all training set sizes. 

Unsurprisingly, the impact of (R-)PERL diminishes as more labeled training data become available: From 7.5\% (R-PERL vs. Fine-tuned BERT) when 100 sentences are available, to 2.1\% for 20K training sentences. To our knowledge, the effectiveness of pivot-based methods for in-domain learning has not been demonstrated in past.


\section{Ablation Analysis and Discussion}
\label{sec:ablation}

In order to shed more light on PERL, we conduct an ablation analysis. We start by uncovering the hyper-parameters that have strong impact on its performance, and analysing its stability across hyper-parameter configurations. We then explore the impact of some of the design choices we made when constructing the model.

In order to keep our analysis concise and to avoid heavy computations, we have to consider only a handful of arbitrarily chosen DA setups for each analysis. We follow the five-fold cross-validation protocol of \S \ref{sec:experiments} for hyper-parameter tuning, except that in some of the analyses a hyper-parameter of interest is kept fixed.



\subsection{Hyper-parameters Analysis}

In this analysis we focus on one hyper-parameter that is relevant only for methods that employ massively pre-trained encoders (the number of unfrozen encoder layers during fine-tuning), as well as on two hyper-parameters that impact the core of our modified MLM objective (number of pivots and the pivot and non-pivot masking probabilities). We finally perform stability analysis across hyper-parameter configurations.

\paragraph{Number of Unfrozen BERT Layers during Fine Tuning (stage 2, Figure~\ref{fig:PERL-fine-tune})}

\begin{figure}[t!]
\includegraphics[width=0.48\textwidth, height=5cm]{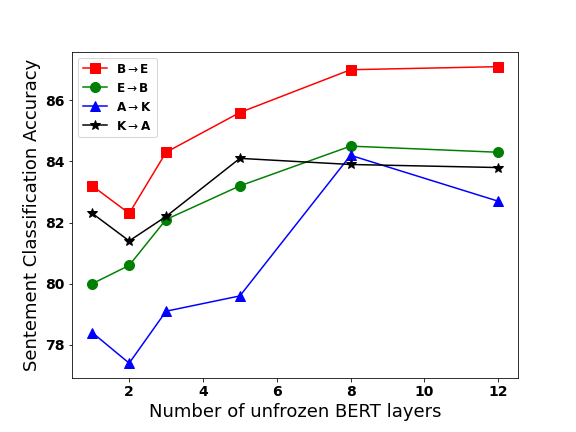}
\caption{The impact of the number of unfrozen PERL layers during fine-tuning (Step 2).}
\label{fig:Num-unfrozen-BERT-layers}
\end{figure}

In Figure~\ref{fig:Num-unfrozen-BERT-layers} we compare PERL final sentiment classification accuracy with six alternatives -- 1, 2, 3, 5, 8 or 12 unfrozen layers, going from the top to the bottom layers. We consider 4 arbitrarily chosen DA setups, where the number of unfrozen layers is kept fixed during the five-fold cross validation process. 
The general trend is clear: PERL performance improves as more layers are unfrozen, and this improvement saturates at 8 unfrozen layers (for the K$\rightarrow$A setup the saturation is at 5 layers). The classification accuracy improvement (compared to 1 unfrozen layer) is of 4\% or more in three of the setups (K$\rightarrow$A is again the exception with only $\sim$ 2\% improvement). Across the experiments of this paper, this hyper-parameter has been the single most influential hyper-parameter of the PERL, R-PERL and Fine-tuned BERT models.

\begin{figure}[t!]
\includegraphics[width=0.48\textwidth]{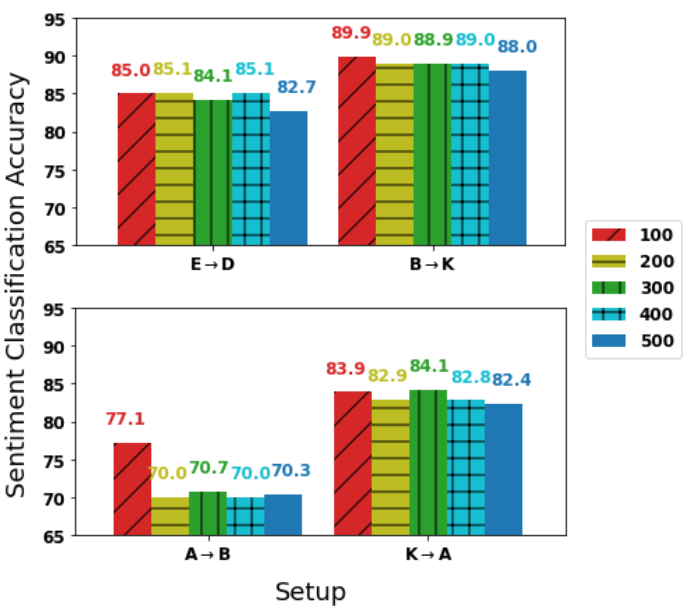}
\caption{PERL sentiment classification accuracy across four setups with a varying number of pivots.}
\label{fig:num_of_pivots}
\end{figure}

\paragraph{Number of Pivots}

Following previous work (e.g. \cite{ziser2018pivot}), our hyper-parameter tuning process considers 100 to 500 pivots in steps of 100. We would next like to explore the impact of this hyper-parameter on PERL performance. 
Figure \ref{fig:num_of_pivots} presents our results, for four arbitrarily selected setups.
In 3 of 4 setups PERL performance is stable across pivot numbers. In 2 setups, 100 is the optimal number of pivots (for the $A \rightarrow B$ setup with a large gap), and in the 2 other setups it lags behind the best value by no more than 0.2\%. These two characteristics -- model stability across pivot numbers and somewhat better performance when using fewer pivots -- were observed across our experiments with PERL and R-PERL. 

\paragraph {Pivot and Non-Pivot Masking Probabilities}

\begin{figure}[t!]
\includegraphics[width=.48\textwidth]{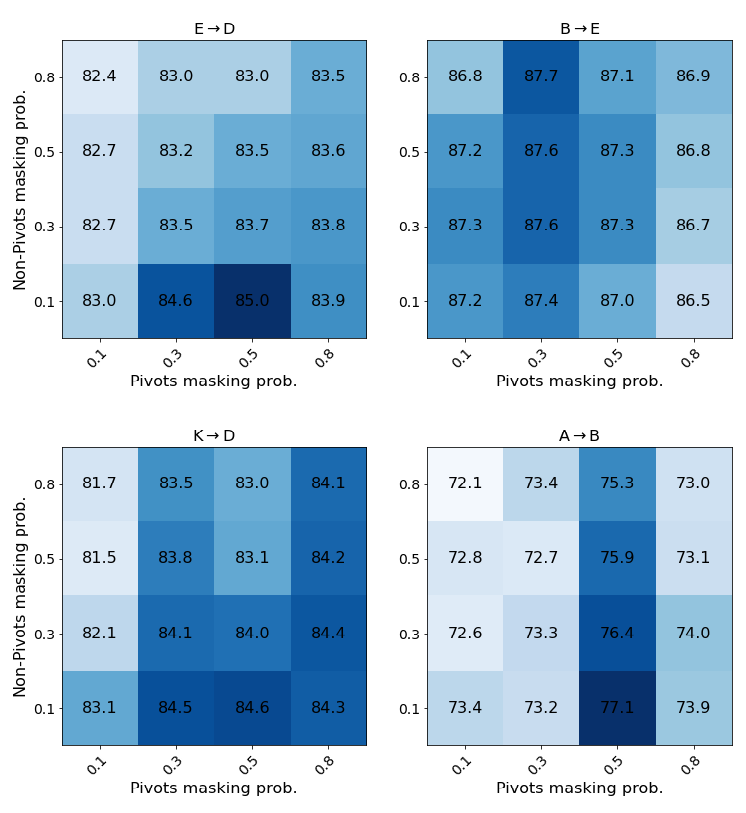}
\caption{Heat maps of PERL performance with different pivot ($\alpha$) and non-pivot ($\beta$) masking probabilities. A darker color corresponds to a higher sentiment classification accuracy.}
\label{fig:Making probs.}
\end{figure}

We next study the impact of the pivot and non-pivot masking probabilities, used during PERL fine-tuning ($\alpha$ and $\beta$, respectively, see \S \ref{sec:DA with PERL}).  For both $\alpha$ and $\beta$ we consider the values of 0.1, 0.3, 0.5 and 0.8. 
%
%
Figure~\ref{fig:Making probs.} presents heat maps that summarize our results. A first observation is the relative stability of PERL to the values of these hyper-parameters: The gap between the best and worst performing configurations are 2.6\% (E $\rightarrow$ D), 1.2\%  (B $\rightarrow$ E), 3.1\% (K $\rightarrow$ D) and 5.0\% (A $\rightarrow$ B). 
A second observation is that extreme $\alpha$ values (0.1 and 0.8) tend to harm the model. Finally, in 3 of 4 cases the best model performance is achieved with $\alpha = 0.5$ and $\beta = 0.1$. 
%

\paragraph{Stability Analysis}

We finally turn to analyse the stability of the PERL models compared to the baselines. Previous work on PBLM and HATN has demonstrated their instability across model configurations (see \citet{ZiserR19} for PBLM and \citet{DBLP:conf/iclr/CuiZSJW19} for HATN). As noted in \citet{ZiserR19}, cross-configuration stability is of particular importance in unsupervised domain adaptation as the hyper-parameter configuration is selected using unlabeled data from the source, rather than the target domain.

In this analysis a hyper-parameter value is not considered for a model if it is not included in the best hyper-parameter configuration of that model for at least one DA setup. Hence, for PERL we fix the number of unfrozen layers (8), the number of pivots (100), and set $(\alpha,\beta) = (0.5,0.1)$, and for PBLM we consider only word embedding size of 128 and 256. Other than that, we consider all possible hyper-parameter configurations of all models (\S \ref{sec:experiments}, 54 configurations for PERL, R-PERL and Fine-tuned BERT, 18 for BERT, 30 for PBLM and 27 for HATN).
Table~\ref{tab:Model stability to configurations.}  presents the minimum
(min), maximum (max), average (avg) and standard deviation (std) of the test set scores across the hyper-parameter configurations of each model, for 4 arbitrarily selected setups. 

In all 4 setups, PERL and R-PERL consistently achieve higher avg, max and min values and lower std values compared to the other models (with the exception of PBLM achieving higher max for $K \rightarrow A$). Moreover, the std values  of PBLM and especially HATN are substantially higher than those of the models that employ BERT. Yet, PERL and R-PERL demonstrate lower std values compared to BERT and Fine-tuned BERT in 3 of 4 setups, indicating that our method contributes to stability beyond the documented contribution of BERT itself \citep{DBLP:conf/emnlp/HaoDWX19}. 


\begin{table}[bt]
\small
\centering
\begin{tabular}{| l | c | c | c | c |}
\hline
\multicolumn{5}{|c|}{\textbf{E}$\boldsymbol{\rightarrow }$\textbf{D}}\\
\hline
& \textbf{avg} & \textbf{max} & \textbf{min}  & \textbf{std} \\
\hline
\textbf{R-PERL} & 84.6 & 85.8 & 83.1 & 0.7 \\
\textbf{PERL} & 85.2 & 86.0 & 84.4 & 0.4 \\
\textbf{Fine-tuned BERT} & 81.3 & 83.2 & 79.0 & 1.2 \\
\textbf{BERT} & 75.0 & 76.8 & 70.6 & 1.8 \\
\textbf{PBLM} & 71.7 & 79.3 & 65.9 & 3.4 \\
\textbf{HATN} & 73.7 & 81.1 & 53.9 & 10.7 \\
\hline
\multicolumn{5}{|c|}{\textbf{B}$\boldsymbol{\rightarrow }$\textbf{K}}\\
\hline
& \textbf{avg} & \textbf{max} & \textbf{min}  & \textbf{std} \\
\hline
\textbf{R-PERL} & 89.5 & 90.5 & 88.8 & 0.5 \\
\textbf{PERL} & 89.4 & 90.2 & 88.8 & 0.3 \\
\textbf{Fine-tuned BERT} & 86.9 & 87.7 & 84.9 & 0.8 \\
\textbf{BERT} & 81.1 & 82.5 & 78.6 & 1.1 \\
\textbf{PBLM} & 78.6 & 84.1 & 71.3 & 3.3 \\
\textbf{HATN} & 76.8 & 82.8 & 59.5 & 7.7 \\
\hline
\multicolumn{5}{|c|}{\textbf{A}$\boldsymbol{\rightarrow }$\textbf{B}}\\
\hline
& \textbf{avg} & \textbf{max} & \textbf{min}  & \textbf{std} \\
\hline
\textbf{R-PERL} & 75.3 & 79.0 & 72.0 & 1.7 \\
\textbf{PERL} & 73.9 & 77.1 & 70.9 & 1.7 \\
\textbf{Fine-tuned BERT} & 72.1 & 74.2 & 68.2 & 1.7 \\
\textbf{BERT} & 69.9 & 73.0 & 66.9 & 1.8 \\
\textbf{PBLM} & 64.2 & 71.6 & 60.9 & 2.7 \\
\textbf{HATN} & 57.6 & 65.0 & 53.7 & 3.5 \\
\hline
\multicolumn{5}{|c|}{\textbf{K}$\boldsymbol{\rightarrow }$\textbf{A}}\\
\hline
& \textbf{avg} & \textbf{max} & \textbf{min}  & \textbf{std} \\
\hline
\textbf{R-PERL} & 85.3 & 86.4 & 84.6 & 0.5 \\
\textbf{PERL} & 83.8 & 84.9 & 81.5 & 0.9 \\
\textbf{Fine-tuned BERT} & 77.8 & 82.1 & 67.1 & 4.2 \\
\textbf{BERT} & 70.4 & 74.0 & 65.1 & 2.6 \\
\textbf{PBLM} & 76.1 & 86.1 & 66.2 & 6.8 \\
\textbf{HATN} & 72.1 & 79.2 & 53.9 & 9.9 \\
\hline
\end{tabular}

\caption{Stability analysis.}
\label{tab:Model stability to configurations.}
\end{table}

\subsection{Design Choice Analysis}

\paragraph{Impact of Pivot Selection}

One design choice that impacts our results is the method through which pivots are selected. 
We next compare three alternatives to our pivot selection method, keeping all other aspects of PERL fixed. As above, we arbitrarily select four setups.

We consider the following pivot selection methods: (a) Random-Frequent: Pivots are randomly selected from the unigrams and bigrams that appear at least 80 times in the unlabeled data of each of the domains; (b) High-MI, No Target: We select the pivots that have the highest mutual information (MI) with the source domain label, but  appear less than 10 times in the target domain unlabeled data; (c) Oracle \citep{oracle_Miller19} : Here the pivots are selected according to our method, but the labeled data used for pivot-label MI computation is the target domain test data rather than the source domain training data. This is an upper bound on the performance of our method since it uses target domain labeled data, which is not available to us. For all methods we select 100 pivots (see above). 

\begin{table*} [t!]
\centering
\begin{adjustbox}{width=\textwidth}
\begin{tabular}{c | c | c| c | c | c | c| c | c }
& \multicolumn{4}{c |}{B $\rightarrow$ E} & \multicolumn{4}{c}{A $\rightarrow$ K} \\
 & 5 layers & 8 layers & 10 layers  & 12 layers (full) & 5 layers & 8 layers & 10 layers  & 12 layers (full)  \\
    \hline
  BERT & 70.9 & 75.9 & 80.6 & 78.8 & 71.2 & 74.9 & 81.2 & 78.8 \\
  Fine-tuned BERT & 74.6 & 76.5 & 84.2 & 84.2 & 74.0 & 76.3 & 80.8 & 81.9 \\
  PERL (Ours) & 81.1 & 83.2 & \textbf{88.2} & 87.0 & 77.7 & 80.2 & \textbf{84.7} & 84.2 \\ 

\end{tabular}
\end{adjustbox}
\caption{Classification accuracy with reduced-size encoders.}
\label{tab:Reduced encoder size.}
\end{table*}

\begin{table}
\centering
\begin{adjustbox}{width=\columnwidth}
\begin{tabular}{c | c | c | c | c}
& B $\rightarrow$ E  & K $\rightarrow$ D & E $\rightarrow$ K & D $\rightarrow$ B\\
    \hline
     BERT & 78.8 & 77.7 & 85.1 & 81.0 \\ 
     Fine-tuned BERT & 84.2 & 79.8 & 89.2 & 84.1 \\ 
     \hline
     High-MI, No Target  & 76.2 & 76.4 & 84.9 & 83.7 \\ 
     Random-Frequent & 79.7 & 76.8 & 85.5 & 81.7 \\ 
     \hline
     PERL (Ours) & \textbf{87.0} & \textbf{84.6} & \textbf{90.6} & \textbf{85.0} \\
     \hline
     \hline
     Oracle & 88.9 & 85.6 & 91.5 & 86.7 \\
\end{tabular}
\end{adjustbox}
\caption{Impact of PERL's pivot selection method.}
\label{tab:Pivots Selection.}
\end{table}

\begin{table}
\centering
\begin{adjustbox}{width=\columnwidth}
\begin{tabular}{c | c | c | c | c}
& B $\rightarrow$ E  & K $\rightarrow$ D & A $\rightarrow$ B & I $\rightarrow$ E \\
    \hline
    \multicolumn{5}{c}{No fine-tuning} \\
    \hline
     BERT & 78.8 & 77.7 & 70.9 & 75.4 \\ 
     \hline
     \multicolumn{5}{c}{Source data only} \\
     \hline
     Fine-tuned BERT & 80.7 & 79.8 & 69.4 & 81.0 \\ 
     PERL  & 79.6 & 82.2 & 69.8 & 84.4 \\
     \hline
     \multicolumn{5}{c}{Target data only} \\
     \hline
     Fine-tuned BERT & 82.0 & 80.9 & 71.6 & 81.1 \\ 
     PERL  & 86.9 & 83.0 & 71.8 & 84.2 \\
     \hline
     \multicolumn{5}{c}{Source and target data} \\
     \hline
     Fine-tuned BERT & 84.2 & 79.8 & 72.9 & 81.5 \\ 
     PERL  & \textbf{87.0} & \textbf{84.6} & \textbf{77.1} & \textbf{87.1} \\
\end{tabular}
\end{adjustbox}
\caption{Impact of fine-tuning data selection.}
\label{tab:unlaveled_data_selection}
\end{table}

Table \ref{tab:Pivots Selection.} presents the results
of the four PERL variants, and compare them to BERT and Fine-tuned BERT. We observe four patterns in the results. First, PERL with our pivot selection method, that emphasizes both high MI with the task label and high frequency in both the source and target domains, is the best performing model. Second, PERL with Random-Frequent pivot selection is substantially outperformed by PERL, but it still performs better than BERT (in 3 of 4 setups), probably because BERT is not tuned on unlabeled data from the participating domains. Yet, PERL with Random-Frequent pivots is outperformed by the Fine-tuned BERT in all setups, indicating that it provides a sub-optimal way of exploiting source and target unlabeled data. Third, in 3 of 4 setups, PERL with the High-MI, No Target pivots is outperformed by the baseline BERT model. This is  a clear indication of the sub-optimality of this pivot selection method which yields a model that is inferior even to a model that was not tuned on source and target domain data. Finally, while, unsurprisingly, PERL with oracle pivots outperforms the standard PERL, the gap is smaller than 2\% in all four cases. Our results clearly demonstrate the strong positive impact of our pivot selection method on the performance of PERL.

\paragraph{Unlabeled Data Selection}

Another design choice we consider is the impact of the type of fine-tuning data. While we followed previous work (e.g. \cite{ziser2018pivot}) and used the unlabeled data from both the source and target domains, it might be that data from only one of the domains, particularly the target, is a better choice. As above, we explore this question on 4 arbitrarily selected domain pairs. The results, presented in Table~\ref{tab:unlaveled_data_selection}, clearly indicate that our choice to use unlabeled data from both domains is optimal, particularly when transferring from a non-product domain (A or I) to a product domain.

\paragraph{Reduced Size Encoder}

We finally explore the effect of the fine-tuning step on the performance of reduced-size models. By doing this we address a major limitation of pre-trained encoders -- their size, which prevents them from running on small computational devices and dictates long run times. 

For this experiment we prune the top encoder layers before its fine-tuning step, yielding three new model sizes, with 5, 8, or 10 layers, compared to the full 12 layers. This is done both for Fine-tuned BERT and for PERL.  We then tune the number of encoder's top unfrozen layers during fine-tuning, as follows: 5 layer-encoder (1, 2, 3); 8 layer-encoder (1, 3, 4, 5); 10 layer-encoder (1, 3, 5, 8); and full encoder (1, 2, 3, 5, 8, 12). For comparison, we employ the BERT model when its top layers are pruned, and no fine-tuning is performed. We focus on two arbitrarily selected DA setups.

Table \ref{tab:Reduced encoder size.} presents accuracy results. In both setups PERL with 10 layers is the best performing model. Moreover, for each number of layers, PERL outperforms the other two models, with particularly substantial improvements for 5 and 8 layers (e.g. 7.3\% and 6.7\%, over BERT and Fine-tuned BERT, respectively, for B $\rightarrow$ E and 8 layers). 

Reduced-size PERL is of course much faster than the full model. The averaged run-time of the full (12 layers) PERL on our test-sets is 196.5 msec and 9.9 msec on CPU (skylake i9-7920X, 2.9 GHz, single thread) and GPU (GeForce GTX 1080 Ti), respectively. For 8 layers the numbers drop to 132.4 msec (CPU) and 6.9 msec (GPU) and for 5 layers to 84.0 (CPU) and 4.7 (GPU) msec.


\section{Conclusions}

We presented PERL, a domain-adaptation model which fine-tunes a massively pre-trained deep contextualized embedding encoder (BERT) with a pivot-based MLM objective. PERL outperforms strong baselines across 22 sentiment classification DA setups, improves in-domain model performance, increases its cross-configuration stability and yields effective reduced-size models.

Our focus in this paper is on binary sentiment classification, as was done in a large body of previous DA work. In future work we would like to extend PERL's reach to structured (e.g. dependency parsing and aspect-based sentiment classification) and generation (e.g. abstractive summarization and machine translation) NLP tasks.

\section*{Acknowledgements}

We would like to thank the action editor and the reviewers, Yftah Ziser, as well as the members of the IE@Technion NLP group for their valuable feedback and advice. This research was partially funded by an ISF personal grant No. 1625/18.

\bibliography{tacl2018}

\begin{thebibliography}{53}
\expandafter\ifx\csname natexlab\endcsname\relax\def\natexlab#1{#1}\fi

\bibitem[{Ben{-}David et~al.(2010)Ben{-}David, Blitzer, Crammer, Kulesza,
  Pereira, and Vaughan}]{ben2010theory}
Shai Ben{-}David, John Blitzer, Koby Crammer, Alex Kulesza, Fernando Pereira,
  and Jennifer~Wortman Vaughan. 2010.
\newblock \href {https://doi.org/10.1007/s10994-009-5152-4} {A theory of
  learning from different domains}.
\newblock \emph{Machine Learning}, 79(1-2):151--175.

\bibitem[{Blitzer et~al.(2007)Blitzer, Dredze, and
  Pereira}]{blitzer2007biographies}
John Blitzer, Mark Dredze, and Fernando Pereira. 2007.
\newblock \href {https://www.aclweb.org/anthology/P07-1056/} {Biographies,
  bollywood, boom-boxes and blenders: Domain adaptation for sentiment
  classification}.
\newblock In \emph{{ACL} 2007, Proceedings of the 45th Annual Meeting of the
  Association for Computational Linguistics, June 23-30, 2007, Prague, Czech
  Republic}. The Association for Computational Linguistics.

\bibitem[{Blitzer et~al.(2006)Blitzer, McDonald, and
  Pereira}]{blitzer2006domain}
John Blitzer, Ryan~T. McDonald, and Fernando Pereira. 2006.
\newblock \href {https://www.aclweb.org/anthology/W06-1615/} {Domain adaptation
  with structural correspondence learning}.
\newblock In \emph{{EMNLP} 2006, Proceedings of the 2006 Conference on
  Empirical Methods in Natural Language Processing, 22-23 July 2006, Sydney,
  Australia}, pages 120--128. {ACL}.

\bibitem[{Bollegala et~al.(2015)Bollegala, Maehara, and
  Kawarabayashi}]{bollegala2015unsupervised}
Danushka Bollegala, Takanori Maehara, and Ken{-}ichi Kawarabayashi. 2015.
\newblock \href {https://doi.org/10.3115/v1/p15-1071} {Unsupervised
  cross-domain word representation learning}.
\newblock In \emph{Proceedings of the 53rd Annual Meeting of the Association
  for Computational Linguistics and the 7th International Joint Conference on
  Natural Language Processing of the Asian Federation of Natural Language
  Processing, {ACL} 2015, July 26-31, 2015, Beijing, China, Volume 1: Long
  Papers}, pages 730--740. The Association for Computer Linguistics.

\bibitem[{Chen et~al.(2011)Chen, Weinberger, and Chen}]{chen2011automatic}
Minmin Chen, Kilian~Q. Weinberger, and Yixin Chen. 2011.
\newblock \href {https://icml.cc/2011/papers/498\_icmlpaper.pdf} {Automatic
  feature decomposition for single view co-training}.
\newblock In \emph{Proceedings of the 28th International Conference on Machine
  Learning, {ICML} 2011, Bellevue, Washington, USA, June 28 - July 2, 2011},
  pages 953--960. Omnipress.

\bibitem[{Chen et~al.(2012)Chen, Xu, Weinberger, and
  Sha}]{chen2012marginalized}
Minmin Chen, Zhixiang~Eddie Xu, Kilian~Q. Weinberger, and Fei Sha. 2012.
\newblock \href {http://icml.cc/2012/papers/416.pdf} {Marginalized denoising
  autoencoders for domain adaptation}.
\newblock In \emph{Proceedings of the 29th International Conference on Machine
  Learning, {ICML} 2012, Edinburgh, Scotland, UK, June 26 - July 1, 2012}.
  icml.cc / Omnipress.

\bibitem[{Clinchant et~al.(2016)Clinchant, Csurka, and
  Chidlovskii}]{clinchant2016domain}
St{\'{e}}phane Clinchant, Gabriela Csurka, and Boris Chidlovskii. 2016.
\newblock \href {https://doi.org/10.18653/v1/p16-2005} {A domain adaptation
  regularization for denoising autoencoders}.
\newblock In \emph{Proceedings of the 54th Annual Meeting of the Association
  for Computational Linguistics, {ACL} 2016, August 7-12, 2016, Berlin,
  Germany, Volume 2: Short Papers}. The Association for Computer Linguistics.

\bibitem[{Cui et~al.(2019)Cui, Zheng, Shen, Jiang, and
  Wang}]{DBLP:conf/iclr/CuiZSJW19}
Wanyun Cui, Guangyu Zheng, Zhiqiang Shen, Sihang Jiang, and Wei Wang. 2019.
\newblock \href {https://openreview.net/forum?id=ByldlhAqYQ} {Transfer learning
  for sequences via learning to collocate}.
\newblock In \emph{7th International Conference on Learning Representations,
  {ICLR} 2019, New Orleans, LA, USA, May 6-9, 2019}. OpenReview.net.

\bibitem[{Devlin et~al.(2019)Devlin, Chang, Lee, and
  Toutanova}]{devlin2018bert}
Jacob Devlin, Ming{-}Wei Chang, Kenton Lee, and Kristina Toutanova. 2019.
\newblock \href {https://doi.org/10.18653/v1/n19-1423} {{BERT:} pre-training of
  deep bidirectional transformers for language understanding}.
\newblock In \emph{Proceedings of the 2019 Conference of the North American
  Chapter of the Association for Computational Linguistics: Human Language
  Technologies, {NAACL-HLT} 2019, Minneapolis, MN, USA, June 2-7, 2019, Volume
  1 (Long and Short Papers)}, pages 4171--4186. Association for Computational
  Linguistics.

\bibitem[{Dozat and Manning(2017)}]{Dozat:17}
Timothy Dozat and Christopher~D. Manning. 2017.
\newblock \href {https://openreview.net/forum?id=Hk95PK9le} {Deep biaffine
  attention for neural dependency parsing}.
\newblock In \emph{5th International Conference on Learning Representations,
  {ICLR} 2017, Toulon, France, April 24-26, 2017, Conference Track
  Proceedings}. OpenReview.net.

\bibitem[{Edunov et~al.(2018)Edunov, Ott, Auli, and Grangier}]{Edunov:18}
Sergey Edunov, Myle Ott, Michael Auli, and David Grangier. 2018.
\newblock \href {https://doi.org/10.18653/v1/d18-1045} {Understanding
  back-translation at scale}.
\newblock In \emph{Proceedings of the 2018 Conference on Empirical Methods in
  Natural Language Processing, Brussels, Belgium, October 31 - November 4,
  2018}, pages 489--500. Association for Computational Linguistics.

\bibitem[{Ganin et~al.(2016)Ganin, Ustinova, Ajakan, Germain, Larochelle,
  Laviolette, Marchand, and Lempitsky}]{Ganin:16}
Yaroslav Ganin, Evgeniya Ustinova, Hana Ajakan, Pascal Germain, Hugo
  Larochelle, Fran{\c{c}}ois Laviolette, Mario Marchand, and Victor~S.
  Lempitsky. 2016.
\newblock \href {http://jmlr.org/papers/v17/15-239.html} {Domain-adversarial
  training of neural networks}.
\newblock \emph{J. Mach. Learn. Res.}, 17:59:1--59:35.

\bibitem[{Glorot et~al.(2011)Glorot, Bordes, and Bengio}]{glorot2011domain}
Xavier Glorot, Antoine Bordes, and Yoshua Bengio. 2011.
\newblock \href {https://icml.cc/2011/papers/342\_icmlpaper.pdf} {Domain
  adaptation for large-scale sentiment classification: {A} deep learning
  approach}.
\newblock In \emph{Proceedings of the 28th International Conference on Machine
  Learning, {ICML} 2011, Bellevue, Washington, USA, June 28 - July 2, 2011},
  pages 513--520. Omnipress.

\bibitem[{Gouws et~al.(2012)Gouws, Van~Rooyen, and Bengio}]{gouws2012learning}
Stephan Gouws, Gert-Jan Van~Rooyen, and Yoshua Bengio. 2012.
\newblock \href
  {https://pdfs.semanticscholar.org/6344/f6e096e8636de0e7943469666f34febae358.pdf}
  {Learning structural correspondences across different linguistic domains with
  synchronous neural language models}.
\newblock In \emph{Proc. of the xLite Workshop on Cross-Lingual Technologies,
  NIPS}.

\bibitem[{Han and Eisenstein(2019)}]{han2019unsupervised}
Xiaochuang Han and Jacob Eisenstein. 2019.
\newblock \href {http://arxiv.org/abs/1904.02817} {Unsupervised domain
  adaptation of contextualized embeddings: {A} case study in early modern
  english}.
\newblock \emph{CoRR}, abs/1904.02817.

\bibitem[{Hao et~al.(2019)Hao, Dong, Wei, and Xu}]{DBLP:conf/emnlp/HaoDWX19}
Yaru Hao, Li~Dong, Furu Wei, and Ke~Xu. 2019.
\newblock \href {https://doi.org/10.18653/v1/D19-1424} {Visualizing and
  understanding the effectiveness of {BERT}}.
\newblock In \emph{Proceedings of the 2019 Conference on Empirical Methods in
  Natural Language Processing and the 9th International Joint Conference on
  Natural Language Processing, {EMNLP-IJCNLP} 2019, Hong Kong, China, November
  3-7, 2019}, pages 4141--4150. Association for Computational Linguistics.

\bibitem[{Huang et~al.(2006)Huang, Smola, Gretton, Borgwardt, and
  Sch{\"{o}}lkopf}]{huang2007correcting}
Jiayuan Huang, Alexander~J. Smola, Arthur Gretton, Karsten~M. Borgwardt, and
  Bernhard Sch{\"{o}}lkopf. 2006.
\newblock \href
  {http://papers.nips.cc/paper/3075-correcting-sample-selection-bias-by-unlabeled-data}
  {Correcting sample selection bias by unlabeled data}.
\newblock In \emph{Advances in Neural Information Processing Systems 19,
  Proceedings of the Twentieth Annual Conference on Neural Information
  Processing Systems, Vancouver, British Columbia, Canada, December 4-7, 2006},
  pages 601--608. {MIT} Press.

\bibitem[{III and Marcu(2006)}]{daume2006domain}
Hal~Daum{\'{e}} III and Daniel Marcu. 2006.
\newblock \href {https://doi.org/10.1613/jair.1872} {Domain adaptation for
  statistical classifiers}.
\newblock \emph{J. Artif. Intell. Res.}, 26:101--126.

\bibitem[{Jiang and Zhai(2007)}]{jiang2007instance}
Jing Jiang and ChengXiang Zhai. 2007.
\newblock \href {https://www.aclweb.org/anthology/P07-1034/} {Instance
  weighting for domain adaptation in {NLP}}.
\newblock In \emph{{ACL} 2007, Proceedings of the 45th Annual Meeting of the
  Association for Computational Linguistics, June 23-30, 2007, Prague, Czech
  Republic}. The Association for Computational Linguistics.

\bibitem[{Lan et~al.(2020)Lan, Chen, Goodman, Gimpel, Sharma, and
  Soricut}]{lan2019albert}
Zhenzhong Lan, Mingda Chen, Sebastian Goodman, Kevin Gimpel, Piyush Sharma, and
  Radu Soricut. 2020.
\newblock \href {https://openreview.net/forum?id=H1eA7AEtvS} {{ALBERT:} {A}
  lite {BERT} for self-supervised learning of language representations}.
\newblock In \emph{8th International Conference on Learning Representations,
  {ICLR} 2020, Addis Ababa, Ethiopia, April 26-30, 2020}. OpenReview.net.

\bibitem[{Lee et~al.(2020)Lee, Yoon, Kim, Kim, Kim, So, and
  Kang}]{lee2019biobert}
Jinhyuk Lee, Wonjin Yoon, Sungdong Kim, Donghyeon Kim, Sunkyu Kim, Chan~Ho So,
  and Jaewoo Kang. 2020.
\newblock \href {https://doi.org/10.1093/bioinformatics/btz682} {Biobert: A
  pre-trained biomedical language representation model for biomedical text
  mining}.
\newblock \emph{Bioinform.}, 36(4):1234--1240.

\bibitem[{Li et~al.(2018)Li, Wei, Zhang, and Yang}]{li2018hierarchical}
Zheng Li, Ying Wei, Yu~Zhang, and Qiang Yang. 2018.
\newblock \href
  {https://www.aaai.org/ocs/index.php/AAAI/AAAI18/paper/view/16873}
  {Hierarchical attention transfer network for cross-domain sentiment
  classification}.
\newblock In \emph{Proceedings of the Thirty-Second {AAAI} Conference on
  Artificial Intelligence, (AAAI-18), the 30th innovative Applications of
  Artificial Intelligence (IAAI-18), and the 8th {AAAI} Symposium on
  Educational Advances in Artificial Intelligence (EAAI-18), New Orleans,
  Louisiana, USA, February 2-7, 2018}, pages 5852--5859. {AAAI} Press.

\bibitem[{Li et~al.(2017)Li, Zhang, Wei, Wu, and
  Yang}]{DBLP:conf/ijcai/LiZWWY17}
Zheng Li, Yu~Zhang, Ying Wei, Yuxiang Wu, and Qiang Yang. 2017.
\newblock \href {https://doi.org/10.24963/ijcai.2017/311} {End-to-end
  adversarial memory network for cross-domain sentiment classification}.
\newblock In \emph{Proceedings of the Twenty-Sixth International Joint
  Conference on Artificial Intelligence, {IJCAI} 2017, Melbourne, Australia,
  August 19-25, 2017}, pages 2237--2243. ijcai.org.

\bibitem[{Liu et~al.(2019)Liu, Ott, Goyal, Du, Joshi, Chen, Levy, Lewis,
  Zettlemoyer, and Stoyanov}]{liu2019roberta}
Yinhan Liu, Myle Ott, Naman Goyal, Jingfei Du, Mandar Joshi, Danqi Chen, Omer
  Levy, Mike Lewis, Luke Zettlemoyer, and Veselin Stoyanov. 2019.
\newblock \href {http://arxiv.org/abs/1907.11692} {Roberta: {A} robustly
  optimized {BERT} pretraining approach}.
\newblock \emph{CoRR}, abs/1907.11692.

\bibitem[{Louizos et~al.(2016)Louizos, Swersky, Li, Welling, and
  Zemel}]{louizos2015variational}
Christos Louizos, Kevin Swersky, Yujia Li, Max Welling, and Richard~S. Zemel.
  2016.
\newblock \href {http://arxiv.org/abs/1511.00830} {The variational fair
  autoencoder}.
\newblock In \emph{4th International Conference on Learning Representations,
  {ICLR} 2016, San Juan, Puerto Rico, May 2-4, 2016, Conference Track
  Proceedings}.

\bibitem[{Maas et~al.(2011)Maas, Daly, Pham, Huang, Ng, and
  Potts}]{maas2011learning}
Andrew~L. Maas, Raymond~E. Daly, Peter~T. Pham, Dan Huang, Andrew~Y. Ng, and
  Christopher Potts. 2011.
\newblock \href {https://www.aclweb.org/anthology/P11-1015/} {Learning word
  vectors for sentiment analysis}.
\newblock In \emph{The 49th Annual Meeting of the Association for Computational
  Linguistics: Human Language Technologies, Proceedings of the Conference,
  19-24 June, 2011, Portland, Oregon, {USA}}, pages 142--150. The Association
  for Computer Linguistics.

\bibitem[{Mansour et~al.(2008)Mansour, Mohri, and
  Rostamizadeh}]{mansour2009domain}
Yishay Mansour, Mehryar Mohri, and Afshin Rostamizadeh. 2008.
\newblock \href
  {http://papers.nips.cc/paper/3550-domain-adaptation-with-multiple-sources}
  {Domain adaptation with multiple sources}.
\newblock In \emph{Advances in Neural Information Processing Systems 21,
  Proceedings of the Twenty-Second Annual Conference on Neural Information
  Processing Systems, Vancouver, British Columbia, Canada, December 8-11,
  2008}, pages 1041--1048. Curran Associates, Inc.

\bibitem[{McClosky et~al.(2010)McClosky, Charniak, and
  Johnson}]{mcclosky2010automatic}
David McClosky, Eugene Charniak, and Mark Johnson. 2010.
\newblock \href {https://www.aclweb.org/anthology/N10-1004/} {Automatic domain
  adaptation for parsing}.
\newblock In \emph{Human Language Technologies: Conference of the North
  American Chapter of the Association of Computational Linguistics,
  Proceedings, June 2-4, 2010, Los Angeles, California, {USA}}, pages 28--36.
  The Association for Computational Linguistics.

\bibitem[{Miller(2019)}]{oracle_Miller19}
Timothy~A. Miller. 2019.
\newblock \href {https://doi.org/10.18653/v1/n19-1039} {Simplified neural
  unsupervised domain adaptation}.
\newblock In \emph{Proceedings of the 2019 Conference of the North American
  Chapter of the Association for Computational Linguistics: Human Language
  Technologies, {NAACL-HLT} 2019, Minneapolis, MN, USA, June 2-7, 2019, Volume
  1 (Long and Short Papers)}, pages 414--419. Association for Computational
  Linguistics.

\bibitem[{Nguyen(2015)}]{Nguyen2015airline}
Quang Nguyen. 2015.
\newblock \href {https://github.com/quankiquanki/skytrax-reviews-dataset} {The
  airline review dataset}.

\bibitem[{Nguyen et~al.(2015)Nguyen, Plank, and
  Grishman}]{DBLP:conf/acl/NguyenPG15}
Thien~Huu Nguyen, Barbara Plank, and Ralph Grishman. 2015.
\newblock \href {https://doi.org/10.3115/v1/p15-1062} {Semantic representations
  for domain adaptation: {A} case study on the tree kernel-based method for
  relation extraction}.
\newblock In \emph{Proceedings of the 53rd Annual Meeting of the Association
  for Computational Linguistics and the 7th International Joint Conference on
  Natural Language Processing of the Asian Federation of Natural Language
  Processing, {ACL} 2015, July 26-31, 2015, Beijing, China, Volume 1: Long
  Papers}, pages 635--644. The Association for Computer Linguistics.

\bibitem[{Pan et~al.(2010)Pan, Ni, Sun, Yang, and Chen}]{pan2010cross}
Sinno~Jialin Pan, Xiaochuan Ni, Jian{-}Tao Sun, Qiang Yang, and Zheng Chen.
  2010.
\newblock \href {https://doi.org/10.1145/1772690.1772767} {Cross-domain
  sentiment classification via spectral feature alignment}.
\newblock In \emph{Proceedings of the 19th International Conference on World
  Wide Web, {WWW} 2010, Raleigh, North Carolina, USA, April 26-30, 2010}, pages
  751--760. {ACM}.

\bibitem[{Peters et~al.(2018)Peters, Neumann, Iyyer, Gardner, Clark, Lee, and
  Zettlemoyer}]{peters2018deep}
Matthew~E. Peters, Mark Neumann, Mohit Iyyer, Matt Gardner, Christopher Clark,
  Kenton Lee, and Luke Zettlemoyer. 2018.
\newblock \href {https://doi.org/10.18653/v1/n18-1202} {Deep contextualized
  word representations}.
\newblock In \emph{Proceedings of the 2018 Conference of the North American
  Chapter of the Association for Computational Linguistics: Human Language
  Technologies, {NAACL-HLT} 2018, New Orleans, Louisiana, USA, June 1-6, 2018,
  Volume 1 (Long Papers)}, pages 2227--2237. Association for Computational
  Linguistics.

\bibitem[{Plank and Moschitti(2013)}]{DBLP:conf/acl/PlankM13}
Barbara Plank and Alessandro Moschitti. 2013.
\newblock \href {https://www.aclweb.org/anthology/P13-1147/} {Embedding
  semantic similarity in tree kernels for domain adaptation of relation
  extraction}.
\newblock In \emph{Proceedings of the 51st Annual Meeting of the Association
  for Computational Linguistics, {ACL} 2013, 4-9 August 2013, Sofia, Bulgaria,
  Volume 1: Long Papers}, pages 1498--1507. The Association for Computer
  Linguistics.

\bibitem[{Radford et~al.(2018)Radford, Narasimhan, Salimans, and
  Sutskever}]{Radford:18}
Alec Radford, Karthik Narasimhan, Tim Salimans, and Ilya Sutskever. 2018.
\newblock \href
  {https://s3-us-west-2.amazonaws.com/openai-assets/research-covers/language-unsupervised/language_understanding_paper.pdf}
  {Improving language understanding by generative pre-training}.
\newblock \emph{URL https://s3-us-west-2. amazonaws.
  com/openai-assets/researchcovers/languageunsupervised/language understanding
  paper. pdf}.

\bibitem[{Radford et~al.(2019)Radford, Wu, Child, Luan, Amodei, and
  Sutskever}]{radford2019language}
Alec Radford, Jeffrey Wu, Rewon Child, David Luan, Dario Amodei, and Ilya
  Sutskever. 2019.
\newblock \href
  {https://www.ceid.upatras.gr/webpages/faculty/zaro/teaching/alg-ds/PRESENTATIONS/PAPERS/2019-Radford-et-al_Language-Models-Are-Unsupervised-Multitask-%20Learners.pdf}
  {Language models are unsupervised multitask learners}.
\newblock \emph{OpenAI Blog}, 1(8).

\bibitem[{Roark and Bacchiani(2003)}]{roark2003supervised}
Brian Roark and Michiel Bacchiani. 2003.
\newblock \href {https://www.aclweb.org/anthology/N03-1027/} {Supervised and
  unsupervised {PCFG} adaptation to novel domains}.
\newblock In \emph{Human Language Technology Conference of the North American
  Chapter of the Association for Computational Linguistics, {HLT-NAACL} 2003,
  Edmonton, Canada, May 27 - June 1, 2003}. The Association for Computational
  Linguistics.

\bibitem[{Rotman and Reichart(2019)}]{rotmandeep}
Guy Rotman and Roi Reichart. 2019.
\newblock \href {https://transacl.org/ojs/index.php/tacl/article/view/1801}
  {Deep contextualized self-training for low resource dependency parsing}.
\newblock \emph{Transactions of the Association for Computational Linguistics},
  7:695--713.

\bibitem[{Rush et~al.(2012)Rush, Reichart, Collins, and
  Globerson}]{rush2012improved}
Alexander~M. Rush, Roi Reichart, Michael Collins, and Amir Globerson. 2012.
\newblock \href {https://www.aclweb.org/anthology/D12-1131/} {Improved parsing
  and {POS} tagging using inter-sentence consistency constraints}.
\newblock In \emph{Proceedings of the 2012 Joint Conference on Empirical
  Methods in Natural Language Processing and Computational Natural Language
  Learning, EMNLP-CoNLL 2012, July 12-14, 2012, Jeju Island, Korea}, pages
  1434--1444. {ACL}.

\bibitem[{Schnabel and Sch{\"{u}}tze(2014)}]{schnabel2014flors}
Tobias Schnabel and Hinrich Sch{\"{u}}tze. 2014.
\newblock \href
  {https://tacl2013.cs.columbia.edu/ojs/index.php/tacl/article/view/183}
  {{FLORS:} fast and simple domain adaptation for part-of-speech tagging}.
\newblock \emph{Transactions of the Association for Computational Linguistics},
  2:15--26.

\bibitem[{Sugiyama et~al.(2007)Sugiyama, Nakajima, Kashima, von B{\"{u}}nau,
  and Kawanabe}]{sugiyama2008direct}
Masashi Sugiyama, Shinichi Nakajima, Hisashi Kashima, Paul von B{\"{u}}nau, and
  Motoaki Kawanabe. 2007.
\newblock \href
  {http://papers.nips.cc/paper/3248-direct-importance-estimation-with-model-selection-and-its-application-to-covariate-shift-adaptation}
  {Direct importance estimation with model selection and its application to
  covariate shift adaptation}.
\newblock In \emph{Advances in Neural Information Processing Systems 20,
  Proceedings of the Twenty-First Annual Conference on Neural Information
  Processing Systems, Vancouver, British Columbia, Canada, December 3-6, 2007},
  pages 1433--1440. Curran Associates, Inc.

\bibitem[{Tu and Wang(2019)}]{DBLP:journals/access/TuW19}
Manshu Tu and Bing Wang. 2019.
\newblock \href {https://doi.org/10.1109/ACCESS.2019.2901929} {Adding prior
  knowledge in hierarchical attention neural network for cross domain sentiment
  classification}.
\newblock \emph{{IEEE} Access}, 7:32578--32588.

\bibitem[{Vincent et~al.(2008)Vincent, Larochelle, Bengio, and
  Manzagol}]{vincent2008extracting}
Pascal Vincent, Hugo Larochelle, Yoshua Bengio, and Pierre{-}Antoine Manzagol.
  2008.
\newblock \href {https://doi.org/10.1145/1390156.1390294} {Extracting and
  composing robust features with denoising autoencoders}.
\newblock In \emph{Machine Learning, Proceedings of the Twenty-Fifth
  International Conference {(ICML} 2008), Helsinki, Finland, June 5-9, 2008},
  volume 307 of \emph{{ACM} International Conference Proceeding Series}, pages
  1096--1103. {ACM}.

\bibitem[{Wolf et~al.(2019)Wolf, Debut, Sanh, Chaumond, Delangue, Moi, Cistac,
  Rault, Louf, Funtowicz, and Brew}]{Wolf2019HuggingFacesTS}
Thomas Wolf, Lysandre Debut, Victor Sanh, Julien Chaumond, Clement Delangue,
  Anthony Moi, Pierric Cistac, Tim Rault, R{\'{e}}mi Louf, Morgan Funtowicz,
  and Jamie Brew. 2019.
\newblock \href {http://arxiv.org/abs/1910.03771} {Huggingface's transformers:
  State-of-the-art natural language processing}.
\newblock \emph{CoRR}, abs/1910.03771.

\bibitem[{Wu et~al.(2016)Wu, Schuster, Chen, Le, Norouzi, Macherey, Krikun,
  Cao, Gao, Macherey, Klingner, Shah, Johnson, Liu, Kaiser, Gouws, Kato, Kudo,
  Kazawa, Stevens, Kurian, Patil, Wang, Young, Smith, Riesa, Rudnick, Vinyals,
  Corrado, Hughes, and Dean}]{DBLP:journals/corr/WuSCLNMKCGMKSJL16}
Yonghui Wu, Mike Schuster, Zhifeng Chen, Quoc~V. Le, Mohammad Norouzi, Wolfgang
  Macherey, Maxim Krikun, Yuan Cao, Qin Gao, Klaus Macherey, Jeff Klingner,
  Apurva Shah, Melvin Johnson, Xiaobing Liu, Lukasz Kaiser, Stephan Gouws,
  Yoshikiyo Kato, Taku Kudo, Hideto Kazawa, Keith Stevens, George Kurian,
  Nishant Patil, Wei Wang, Cliff Young, Jason Smith, Jason Riesa, Alex Rudnick,
  Oriol Vinyals, Greg Corrado, Macduff Hughes, and Jeffrey Dean. 2016.
\newblock \href {http://arxiv.org/abs/1609.08144} {Google's neural machine
  translation system: Bridging the gap between human and machine translation}.
\newblock \emph{CoRR}, abs/1609.08144.

\bibitem[{Xue et~al.(2008)Xue, Dai, Yang, and Yu}]{xue2008topic}
Gui{-}Rong Xue, Wenyuan Dai, Qiang Yang, and Yong Yu. 2008.
\newblock \href {https://doi.org/10.1145/1390334.1390441} {Topic-bridged {PLSA}
  for cross-domain text classification}.
\newblock In \emph{Proceedings of the 31st Annual International {ACM} {SIGIR}
  Conference on Research and Development in Information Retrieval, {SIGIR}
  2008, Singapore, July 20-24, 2008}, pages 627--634. {ACM}.

\bibitem[{Yang and Eisenstein(2014)}]{yang2014fast}
Yi~Yang and Jacob Eisenstein. 2014.
\newblock \href {https://doi.org/10.3115/v1/p14-2088} {Fast easy unsupervised
  domain adaptation with marginalized structured dropout}.
\newblock In \emph{Proceedings of the 52nd Annual Meeting of the Association
  for Computational Linguistics, {ACL} 2014, June 22-27, 2014, Baltimore, MD,
  USA, Volume 2: Short Papers}, pages 538--544. The Association for Computer
  Linguistics.

\bibitem[{Yang et~al.(2019)Yang, Dai, Yang, Carbonell, Salakhutdinov, and
  Le}]{yang2019xlnet}
Zhilin Yang, Zihang Dai, Yiming Yang, Jaime~G. Carbonell, Ruslan Salakhutdinov,
  and Quoc~V. Le. 2019.
\newblock \href
  {http://papers.nips.cc/paper/8812-xlnet-generalized-autoregressive-pretraining-for-language-understanding}
  {Xlnet: Generalized autoregressive pretraining for language understanding}.
\newblock In \emph{Advances in Neural Information Processing Systems 32: Annual
  Conference on Neural Information Processing Systems 2019, NeurIPS 2019, 8-14
  December 2019, Vancouver, BC, Canada}, pages 5754--5764.

\bibitem[{Yu and Jiang(2016)}]{yu2016learning}
Jianfei Yu and Jing Jiang. 2016.
\newblock \href {https://doi.org/10.18653/v1/d16-1023} {Learning sentence
  embeddings with auxiliary tasks for cross-domain sentiment classification}.
\newblock In \emph{Proceedings of the 2016 Conference on Empirical Methods in
  Natural Language Processing, {EMNLP} 2016, Austin, Texas, USA, November 1-4,
  2016}, pages 236--246. The Association for Computational Linguistics.

\bibitem[{Zhang et~al.(2019)Zhang, Han, Liu, Jiang, Sun, and
  Liu}]{zhang2019ernie}
Zhengyan Zhang, Xu~Han, Zhiyuan Liu, Xin Jiang, Maosong Sun, and Qun Liu. 2019.
\newblock \href {https://doi.org/10.18653/v1/p19-1139} {{ERNIE:} enhanced
  language representation with informative entities}.
\newblock In \emph{Proceedings of the 57th Conference of the Association for
  Computational Linguistics, {ACL} 2019, Florence, Italy, July 28- August 2,
  2019, Volume 1: Long Papers}, pages 1441--1451. Association for Computational
  Linguistics.

\bibitem[{Ziser and Reichart(2017)}]{ziser2016neural}
Yftah Ziser and Roi Reichart. 2017.
\newblock \href {https://doi.org/10.18653/v1/K17-1040} {Neural structural
  correspondence learning for domain adaptation}.
\newblock In \emph{Proceedings of the 21st Conference on Computational Natural
  Language Learning (CoNLL 2017), Vancouver, Canada, August 3-4, 2017}, pages
  400--410. Association for Computational Linguistics.

\bibitem[{Ziser and Reichart(2018)}]{ziser2018pivot}
Yftah Ziser and Roi Reichart. 2018.
\newblock \href {https://doi.org/10.18653/v1/n18-1112} {Pivot based language
  modeling for improved neural domain adaptation}.
\newblock In \emph{Proceedings of the 2018 Conference of the North American
  Chapter of the Association for Computational Linguistics: Human Language
  Technologies, {NAACL-HLT} 2018, New Orleans, Louisiana, USA, June 1-6, 2018,
  Volume 1 (Long Papers)}, pages 1241--1251. Association for Computational
  Linguistics.

\bibitem[{Ziser and Reichart(2019)}]{ZiserR19}
Yftah Ziser and Roi Reichart. 2019.
\newblock \href {https://doi.org/10.18653/v1/p19-1591} {Task refinement
  learning for improved accuracy and stability of unsupervised domain
  adaptation}.
\newblock In \emph{Proceedings of the 57th Conference of the Association for
  Computational Linguistics, {ACL} 2019, Florence, Italy, July 28- August 2,
  2019, Volume 1: Long Papers}, pages 5895--5906. Association for Computational
  Linguistics.

\end{thebibliography}
\bibliographystyle{acl_natbib}
\end{document}